\documentclass[10pt]{article} 
\usepackage[preprint]{tmlr}

\usepackage{amsmath,amsfonts,bm}

\def\eqref#1{equation~\ref{#1}}

\def\1{\bm{1}}

\DeclareMathAlphabet{\mathsfit}{\encodingdefault}{\sfdefault}{m}{sl}
\SetMathAlphabet{\mathsfit}{bold}{\encodingdefault}{\sfdefault}{bx}{n}

\usepackage{hyperref}
\usepackage{url}
\usepackage{graphicx}
\usepackage{subcaption}
\usepackage{amssymb}
\usepackage{xcolor}
\usepackage{longtable}
\newcommand{\best}[1]{\textcolor{red}{\textbf{#1}}}
\newcommand{\second}[1]{\underline{#1}}
\title{Towards Principled Test-Time Adaptation for Time Series Forecasting}

\author{\name Haochun Wang \email haochun.wang@stonybrook.edu \\
      \addr Stony Brook University
      \AND
      \name Ruichen Xu \email ruichen.xu@stonybrook.edu \\
      \addr Stony Brook University
      \AND
      \name Georgios Kementzidis \email georgios.kementzidis@stonybrook.edu\\
      \addr Stony Brook University
      \AND
      \name Karen Cho \email karen.cho@stonybrook.edu\\
      \addr Stony Brook University
      \AND
      \name Sebastian Ramirez Villarreal \email sebastian.ramirezvillarreal@stonybrook.edu\\
      \addr Stony Brook University
      \AND
      \name Yuefan Deng \email yuefan.deng@stonybrook.edu\\
      \addr Stony Brook University}

\def\month{MM}  
\def\year{YYYY} 
\def\openreview{\url{https://openreview.net/forum?id=XXXX}} 

\begin{document}

\maketitle

\begin{abstract}
Test-time adaptation (TTA) has recently emerged as a promising approach for improving time series forecasting (TSF) under distribution shift. Existing TSF-TTA methods differ in how they utilize revealed targets, yet the resulting adaptation protocols remain heterogeneous and lack a clearly unified formulation. To address this issue, we revisit TSF-TTA from the perspective of protocol cleanliness and propose an adaptation protocol based solely on matured ground truth, yielding a more principled setting for adaptation. Under this protocol, we further diagnose existing adapters in the frequency domain and find that their prediction corrections often exhibit limited and weakly structured spectral modifications. Motivated by this diagnosis, we propose Frequency-Aware Calibration (FAC), a lightweight calibration method that directly parameterizes prediction corrections in the frequency domain. Across diverse datasets, forecasting horizons, and source forecasters, FAC achieves competitive and consistent performance while requiring substantially fewer trainable parameters than the compared TSF-TTA adapters.
\end{abstract}

\section{Introduction}
Time series forecasting (TSF) plays a critical role in a wide range of domains, including weather \citep{Wu2023, verma2024climode}, energy \citep{MALEKI2024}, traffic \citep{Yin2022, Zhang2024}, and finance \citep{tsay2010analysis, koa2026reasoning}. Recent advances in deep learning-based forecasting models have substantially improved predictive performance across diverse time series forecasting settings \citep{Zeng2023DLinear,nie2023a,liu2024itransformer}.

Despite this progress, deploying forecasters to real-world time series remains challenging due to the complex and evolving nature of temporal data. In particular, distribution shifts induced by non-stationarity pose a major obstacle to reliable forecasting \citep{granger2003time, Du2021Ada}, especially in long-horizon settings. To address this issue, prior studies have mainly followed two directions: introducing learnable normalization modules \citep{kim2022revin, Fan2023, liu2023adaptive}, which primarily operate during pre-training, or training forecasting models in an online manner \citep{pham2023learning, zhang2023onenet, liang2024actnow, lau2025fast, huang2026online}. While these approaches can partially mitigate the effects of distribution shift, they either focus largely on the training stage or require training an online forecaster from scratch. In light of these limitations, applying test-time adaptation (TTA) \citep{HyunGi2025, medeiros2025accurate, grover2025shiftaware, im2026cosa} to a source forecaster has emerged as a promising alternative. 

\begin{figure}[t]
    \centering
    \begin{subfigure}[t]{0.92\linewidth}
        \centering
        \includegraphics[width=\linewidth]{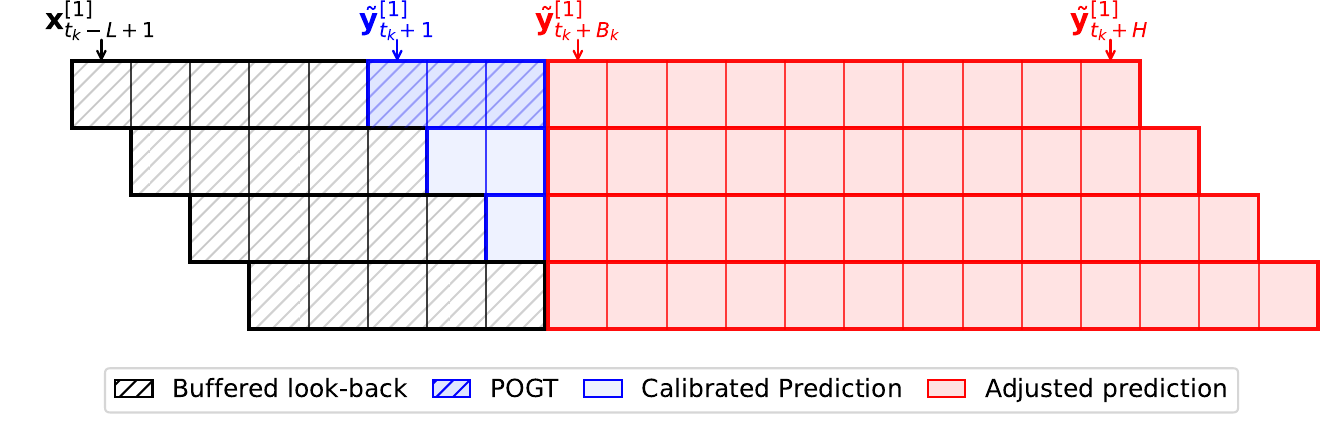}
        \caption{}
        \label{fig:tafas_protocol}
    \end{subfigure}

    \begin{subfigure}[t]{0.92\linewidth}
        \centering
        \includegraphics[width=\linewidth]{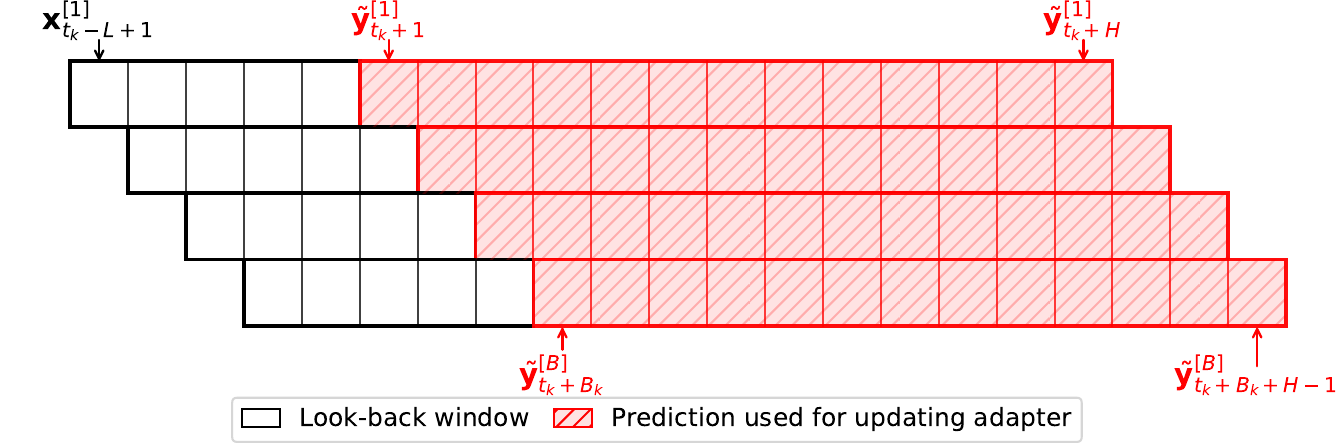}
        \caption{}
        \label{fig:cosa_protocol}
    \end{subfigure}
    \caption{Comparison of adaptation supervision protocols in TSF-TTA. (a) Mixed supervision uses POGT from the current mini-batch, shown in blue, together with matured ground truth from a past mini-batch; the latter is omitted for visual clarity. (b) Streaming adaptation updates the adapter using revealed targets associated with the current mini-batch, indicated by the red shaded region.}
    \vspace{0.6em}
    \label{fig:protocol_comparison}
\end{figure}

Existing TSF-TTA methods differ in how they utilize revealed targets, with some \citep{HyunGi2025, medeiros2025accurate, grover2025shiftaware} incorporating partially-observed ground truth (POGT) together with full ground truth, which we refer to as \textbf{matured ground truth} throughout this paper, to enable more proactive adaptation. At the mini-batch level, a past mini-batch is regarded as matured when the full target horizon for every sample in that mini-batch has been completely observed. Intuitively, with respect to the current batch, matured supervision lies entirely to the left of the end of the first look-back window in the current mini-batch. In contrast, COSA \citep{im2026cosa} is presented under a streaming formulation based on revealed ground truth, where an output adapter is updated on the current mini-batch using targets revealed for that mini-batch and lightweight context constructed from previously observed targets, without being explicitly framed in terms of matured ground truth. As illustrated in Figure~\ref{fig:protocol_comparison}, these two families differ in when revealed targets are consumed for adaptation.

Despite these design differences, the protocol-level role of revealed targets in TSF-TTA remains insufficiently clarified. Our analysis suggests that existing TSF-TTA protocols have not yet provided a clearly unified account of how revealed targets should be used for adaptation. This leads us to revisit TSF-TTA from the perspective of protocol cleanliness. In particular, we advocate using matured ground truth as a cleaner and more principled adaptation signal in rolling forecasting settings. We further examine the prediction corrections of existing TSF-TTA methods in the frequency domain, finding that they often exhibit relatively smooth and weakly structured spectral patterns despite their different calibration designs. Based on this observation, we propose Frequency-Aware Calibration (FAC), which achieves competitive and consistent performance under the proposed protocol across diverse forecasting settings.

Our contributions are summarized as follows:
\begin{itemize}
    \item We revisit how revealed targets are used in TSF-TTA and show that a cleaner protocol based on matured ground truth provides a more principled setting for adaptation.

    \item We provide a frequency-domain diagnosis of existing TSF-TTA methods, showing that their realized corrections often exhibit limited and weakly structured spectral modifications under rolling adaptation.

    \item We propose Frequency-Aware Calibration (FAC), a lightweight adapter that explicitly models structured spectral residuals in the frequency domain and achieves competitive and consistent performance across diverse forecasting settings.
\end{itemize}

\section{Related Work}
\subsection{Time Series Forecasting under Distribution Shift}
A representative line of work addresses distribution shift in time series forecasting through learnable normalization or related distribution calibration mechanisms. RevIN \citep{kim2022revin} mitigates temporal distribution shift via reversible instance normalization and denormalization with learnable affine transformation. Dish-TS \citep{Fan2023} characterizes both intra-space shift and inter-space shift, alleviating them through dual distribution modeling for the input and output spaces. SAN \citep{liu2023adaptive} further models non-stationarity by capturing statistical variation at the temporal-slice level. Although effective, these methods primarily improve robustness through training-time or architecture-level treatment of non-stationarity, rather than adapting a deployed source forecaster at test time. 

Another line of work \citep{pham2023learning, zhang2023onenet, liang2024actnow, lau2025fast, huang2026online} considers online time series forecasting (OTSF), where the forecasting model is continuously updated as new observations arrive. Recent studies \citep{liang2024actnow, lau2025fast} have pointed out that conventional OTSF protocols may suffer from evaluation leakage or unrealistic forecasting setups. In particular, Act-Now \citep{liang2024actnow} and DSOF \citep{lau2025fast} revisit online forecasting protocols and advocate cleaner evaluation settings that avoid evaluating predictions on steps already used for model updates. Within this line of work, Act-Now \citep{liang2024actnow} introduces a label decomposition model to tackle distribution shift, while DSOF \citep{lau2025fast} adopts a dual-stream teacher-student framework for online forecasting. ADAPT-Z \citep{huang2026online} further studies the challenge of delayed-feedback online prediction and argues that distribution shift may be better addressed through feature-level adjustment. These approaches are related to TSF-TTA in spirit, but the primary objectives differ: OTSF focuses on training the online forecaster itself, while TSF-TTA adapts a deployed frozen source forecaster at test time using adaptation modules.

\subsection{Test-Time Adaptation for Time Series Forecasting}
Test-time adaptation (TTA) \citep{wang2021tent, Wang2022, gong2023sotta, Liang2025, kim2026buffer} has been widely studied in domains such as computer vision, where models are adapted at test time to improve robustness under distribution shift. Unlike conventional TTA methods \citep{wang2021tent, Wang2022, gong2023sotta, kim2026buffer}, which typically rely on unlabeled test-time data, TSF-TTA can exploit forecasting targets that become sequentially revealed after prediction. As the pioneering work on TSF-TTA, TAFAS \citep{HyunGi2025} incorporates both matured ground truth and partially observed ground-truth (POGT), aiming to enable earlier and more proactive adaptation. PETSA \citep{medeiros2025accurate} further improves parameter efficiency by introducing low-rank adapters together with a specialized adaptation loss. DynaTTA \citep{grover2025shiftaware} employs dynamic adaptation rates and shift-conditioned gating based on real-time estimation of distribution shift. In contrast, COSA \citep{im2026cosa} is presented under a streaming formulation based on revealed ground truth, where a single output adapter corrects the current mini-batch prediction using lightweight context from previously revealed targets and is updated directly against the targets of the current mini-batch. These methods therefore differ not only in adapter design, but also in when and how revealed targets are incorporated into adaptation.

\section{Towards Cleaner Adaptation Protocols for TSF-TTA}

\subsection{Preliminaries: Rolling TSF-TTA Formulation}
Existing TSF-TTA \citep{HyunGi2025, medeiros2025accurate, grover2025shiftaware, im2026cosa} methods adopt a rolling test-time adaptation setting, in which a pre-trained source forecaster is deployed on temporally ordered test data under distribution shift. At test time, rolling windows are grouped into mini-batches, and predictions are produced for all samples within each mini-batch. Let $L$ and $H$ denote the look-back length and forecasting horizon, respectively. We index rolling mini-batches by $k$, and let $B_k$ denote the size of the $k$-th mini-batch. We further let $t_k$ denote the global time index immediately preceding the first forecasted target of the $k$-th mini-batch. Since rolling windows are grouped consecutively, the next mini-batch starts after $B_k$ shifts, yielding $t_{k+1} = t_k + B_k$. For the $j$-th sample of the $k$-th mini-batch, where $j \in \{1, \dots, B_k\}$, the input window and the corresponding forecast are given by
\begin{equation}
\mathbf{X}^{[j]}_{k} = [\mathbf{x}_{t_k+j-L}, \dots, \mathbf{x}_{t_k+j-1}], \quad \hat{\mathbf{Y}}^{[j]}_{k} = [\hat{\mathbf{y}}^{[j]}_{t_k+j}, \dots, \hat{\mathbf{y}}^{[j]}_{t_k+j+H-1}]
\end{equation}
where $\mathbf{x}_t \in \mathbb{R}^C$ denotes the multivariate observation at time step $t$, and $\hat{\mathbf{y}}^{[j]}_t \in \mathbb{R}^C$ denotes the prediction of the $j$-th sample at time step $t$. Note that $\hat{\mathbf{y}}^{[j]}_t$ differs for different samples in the mini-batch. 

Denote the corresponding target vector as
\begin{equation}
{\mathbf{Y}}^{[j]}_{k} = [{\mathbf{y}}_{t_k+j}, \dots, {\mathbf{y}}_{t_k+j+H-1}],
\end{equation}
then we can denote the forecasting target span of the $k$-th mini-batch as the union of all sample-level targets within the mini-batch, which is given by
\begin{equation}
\mathfrak{Y}_k = \bigcup_{j=1}^{B_k} \mathbf{Y}^{[j]}_{k} = \{\mathbf{y}_{t_k+1}, \dots, \mathbf{y}_{t_k+H+B_k-1}\}.
\end{equation}
\subsection{A Closer Look at Existing Supervision Protocols}
\subsubsection{Mixed Supervision with POGT and Matured Ground Truth}
A representative line of TSF-TTA methods \citep{HyunGi2025, medeiros2025accurate, grover2025shiftaware} adopts a mixed supervision protocol that combines POGT from the current mini-batch with supervision from the most recent matured mini-batch, as illustrated in Figure~\ref{fig:tafas_protocol}. In these methods, the POGT length $p_k$ is determined by the periodicity-aware adaptation scheduling (PAAS), with $p_k = B_k-1.$ For notational consistency, we will use $B_k-1$ to denote the POGT length below. 

Let $\mathcal{C}_{k-1}^{\mathrm{in}}(\cdot)$ and $\mathcal{C}_{k-1}^{\mathrm{out}}(\cdot)$ denote the input and output calibration modules available before adapting on the $k$-th mini-batch. For the $j$-th sample in that mini-batch, the calibrated prediction is 
\begin{equation}
\hat{\mathbf{Y}}^{\mathrm{cali}, [j]}_k = \mathcal{C}_{k-1}^{\mathrm{out}}(\mathcal{F}(\mathcal{C}_{k-1}^{\mathrm{in}}(\mathbf{X}^{[j]}_k))) = [\hat{\mathbf{y}}^{\mathrm{cali}, [j]}_{t_k+j}, \dots, \hat{\mathbf{y}}^{\mathrm{cali}, [j]}_{t_k+j+H-1}],
\end{equation}
where $\mathcal{F}(\cdot)$ denotes the frozen source forecaster. Let $m(k)$ denote the index of the most recent matured mini-batch. Then the mixed supervision protocol can be abstractly written as
\begin{equation}
\mathcal{L}^{\mathrm{mixed}}_k
=
\mathcal{L}^{\mathrm{POGT}}_k + \mathcal{L}^{\mathrm{Matured}}_{m(k)},
\end{equation}
with
\begin{equation}
\mathcal{L}^{\mathrm{POGT}}_k
=
\mathrm{MSE}\!\left(
\{\hat{\mathbf{y}}^{\mathrm{cali},[1]}_{t}\}_{t=t_k+1}^{t_k+B_k-1},
\{\mathbf{y}_{t}\}_{t=t_k+1}^{t_k+B_k-1}
\right).
\end{equation}
and
\begin{equation}
\mathcal{L}^{\mathrm{Matured}}_{m(k)}
=
\mathrm{MSE}\!\left(
\{\hat{\mathbf{Y}}^{\mathrm{cali},[j]}_{m(k)}\}_{j=1}^{B_{m(k)}},
\{\mathbf{Y}^{[j]}_{m(k)}\}_{j=1}^{B_{m(k)}}
\right).
\end{equation}
In implementation, these two terms may be consumed in separate adaptation steps, but together they define the mixed supervision protocol. After adaptation, the calibration modules are updated to $\mathcal{C}_{k}^{\mathrm{in}}(\cdot)$ and $\mathcal{C}_{k}^{\mathrm{out}}(\cdot)$. The current mini-batch is then re-forecast using the updated calibration modules: 
\begin{equation}
\hat{\mathbf{Y}}^{\mathrm{adp},[j]}_k
=
\mathcal{C}_{k}^{\mathrm{out}}
\!\left(
\mathcal{F}\!\left(
\mathcal{C}_{k}^{\mathrm{in}}(\mathbf{X}^{[j]}_k)
\right)
\right).
\end{equation}
The adjusted prediction keeps the already observed prefix from the original calibrated prediction and replaces the unobserved suffix with the re-forecasted one, which is given by
\begin{equation}
\tilde{\mathbf{y}}^{[j]}_{k,t}
=
\begin{cases}
\hat{\mathbf{y}}^{\mathrm{cali},[j]}_{t}, 
& t_k+j \le t \le t_k+B_k-1,\\
\hat{\mathbf{y}}^{\mathrm{adp},[j]}_{t}, 
& t_k+B_k \le t \le t_k+j+H-1.
\end{cases}
\end{equation}

\begin{figure}[htbp]
    \centering
    \begin{subfigure}[t]{\linewidth}
        \centering
        \includegraphics[width=\linewidth]{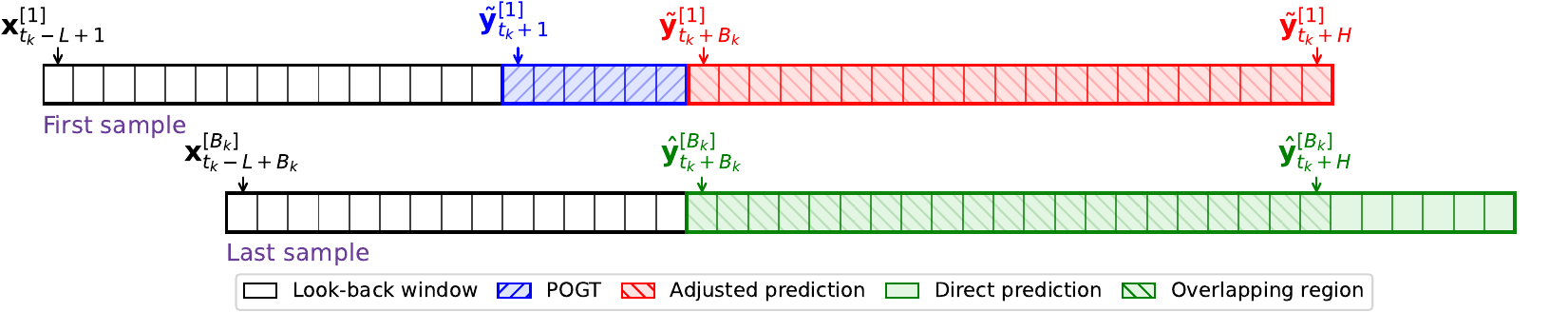}
        \caption{}
        \label{fig:early_vs_late}
    \end{subfigure}

    \begin{subfigure}[t]{\linewidth}
        \centering
        \includegraphics[width=\linewidth]{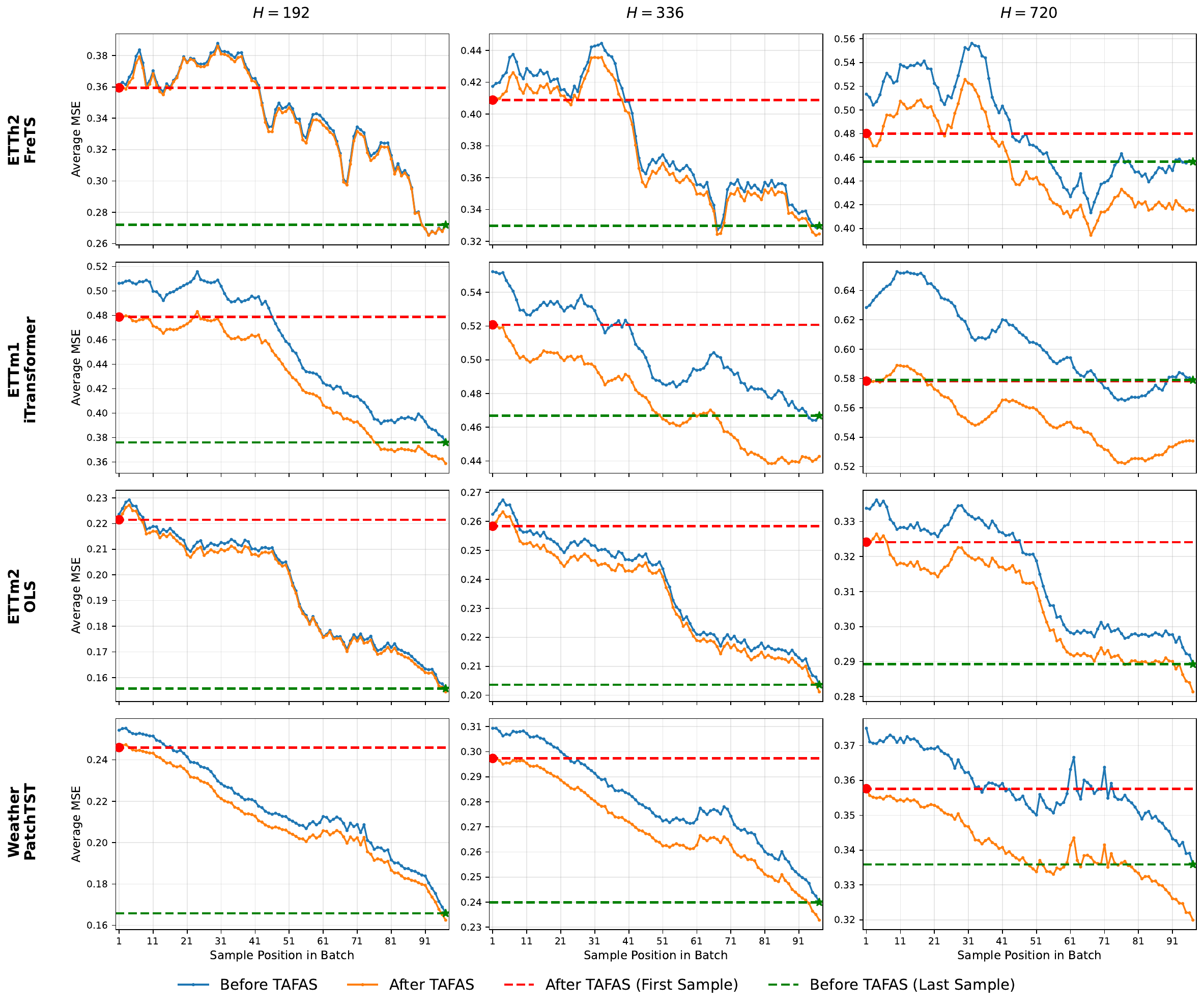}
        \caption{}
        \label{fig:first_vs_last}
    \end{subfigure}
    \caption{Comparison between the adjusted prediction of the first sample and later direct predictions within the same mini-batch. (a) Schematic illustration of the overlapping horizon shared by the first sample's adjusted prediction and the last sample's direct prediction. (b) Mean overlapping-region MSE at each sample position, averaged over mini-batches with $B_k=97$ across representative datasets, horizons, and source forecasters. The blue and orange curves denote direct and adjusted predictions, respectively; the red dashed line marks the adjusted prediction of the first sample, and the green dashed line marks the direct prediction of the last sample. Additional results for $B_k=25$ are provided in Appendix~\ref{app:early_late_additional}.}
    \label{fig:early_late_comparisons}
\end{figure}

However, by the time POGT becomes available for adaptation, the same revealed values have already entered the look-back windows of later samples within the same mini-batch. \textbf{This raises a natural question: is it more beneficial to use these revealed values as supervision for adaptation, or simply as additional input context for later direct prediction?} To examine this, we compare the MSE on the shared overlapping horizon between the adjusted prediction of the first sample and the direct predictions of later samples within the same mini-batch.

As illustrated in Figure~\ref{fig:early_vs_late}, the endpoint comparison provides the most direct way to assess whether using the revealed values as supervision offers a meaningful advantage over simply using the same values as input context in later direct prediction. At the same time, this endpoint view is only the most intuitive illustration of a broader phenomenon within the mini-batch. To examine this more comprehensively, we compute the mean overlapping-region MSE at each sample position over the mini-batches with $B_k=97$, since this is the most common mini-batch size induced by PAAS in our experiments. 

Let $\mathcal{I}_{\{ B_k = B\}}$ denote the set of mini-batches satisfying $B_k = B$. We define the mean overlapping-region MSE for the direct prediction of the $j$-th sample as
\begin{equation}
\hat{M}_{\mathcal{I}_{\{B_k=B\}}}^{[j]}
=
\frac{1}{\left|\mathcal{I}_{\{B_k=B\}}\right|}
\sum_{k \in \mathcal{I}_{\{B_k=B\}}}
\mathrm{MSE}
\bigl(
[\hat{\mathbf{y}}^{[j]}_{t_k+B}, \dots, \hat{\mathbf{y}}^{[j]}_{t_k+H}],
[\mathbf{y}_{t_k+B}, \dots, \mathbf{y}_{t_k+H}]
\bigr),
\end{equation}
and similarly, the mean overlapping-region MSE for the adjusted prediction of the $j$-th sample as
\begin{equation}
\tilde{M}_{\mathcal{I}_{\{B_k=B\}}}^{[j]}
=
\frac{1}{\left|\mathcal{I}_{\{B_k=B\}}\right|}
\sum_{k \in \mathcal{I}_{\{B_k=B\}}}
\mathrm{MSE}
\bigl(
[\tilde{\mathbf{y}}^{[j]}_{t_k+B}, \dots, \tilde{\mathbf{y}}^{[j]}_{t_k+H}],
[\mathbf{y}_{t_k+B}, \dots, \mathbf{y}_{t_k+H}]
\bigr).
\end{equation}
Figure~\ref{fig:first_vs_last} further shows that this phenomenon is more widespread. Across multiple datasets, forecasting horizons and frozen source forecasters, the adjusted prediction of the first sample is generally inferior to the direct prediction of the last sample. In fact, the direct predictions from many intermediate sample positions already achieve comparable or lower overlapping-region MSE than the adjusted prediction of the first sample. These results suggest that using revealed values as supervision is not consistently more beneficial than simply using the same values as input context for later direct prediction.

\subsubsection{Streaming Adaptation with Revealed Ground Truth}
COSA \citep{im2026cosa} adopts a streaming adaptation protocol based on revealed ground truth, as illustrated in Figure~\ref{fig:cosa_protocol}. Under this protocol, the current mini-batch is first predicted by the frozen source forecaster, after which an output adapter produces the final adapter-corrected prediction using a lightweight context vector constructed from previously revealed targets. Unlike the mixed-supervision protocol above, the output adapter is then updated directly by minimizing the MSE loss between these adapter-corrected predictions of the current mini-batch and the revealed ground truth of the current mini-batch.

Let $\mathcal{A}_{k-1}(\cdot,\cdot)$ denote the output adapter available before adapting on the $k$-th mini-batch, and let $\mathbf{c}_{k-1}$ denote the context vector constructed from previously revealed targets. For the $j$-th sample in the $k$-th mini-batch, the final streaming prediction is given by 
\begin{equation}
\hat{\mathbf{Y}}^{\mathrm{stream}, [j]}_k = \mathcal{A}_{k-1}(\mathcal{F}(\mathbf{X}^{[j]}_k), \mathbf{c}_{k-1}). 
\end{equation}
Accordingly, the streaming adaptation objective can be written as 
\begin{equation}
\mathcal{L}_k^{\mathrm{stream}} = \frac{1}{B_k} \sum_{j=1}^{B_k} \mathrm{MSE}(\hat{\mathbf{Y}}^{\mathrm{stream}, [j]}_k, \mathbf{Y}_k^{[j]}).
\end{equation}

Although the protocol is presented in terms of revealed ground truth, such supervision is not necessarily matured relative to subsequent rolling predictions. Recall that the target span of the $k$-th mini-batch is $\mathfrak{Y}_k = \{\mathbf{y}_{t_k+1}, \dots, \mathbf{y}_{t_k+H+B_k-1}\}$. For the $(k+1)$-th mini-batch, since $t_{k+1}=t_k+B_k$, the prediction span of the $j$-th sample is
\begin{equation}
\hat{\mathbf{Y}}^{\mathrm{stream},[j]}_{k+1}
=
[\hat{\mathbf{y}}^{\mathrm{stream},[j]}_{t_k+B_k+j}, \dots, \hat{\mathbf{y}}^{\mathrm{stream},[j]}_{t_k+B_k+j+H-1}].
\end{equation}
When the forecasting horizon is sufficiently larger than the mini-batch size, the targets used to update on batch $k$ may still overlap with the future prediction span of later mini-batches. For example, if $H \ge B_k + B_{k+1}$, then
\begin{equation}
\{t_k+1, \dots, t_k+H+B_k-1\}
\cap
\{t_k+B_k+j, \dots, t_k+B_k+j+H-1\}
=
\{t_k+B_k+j, \dots, t_k+H+B_k-1\}
\neq \emptyset.
\end{equation}
This shows that the supervision used to update on batch $k$ is not necessarily matured relative to subsequent rolling prediction spans.

\subsection{A Cleaner Adaptation Protocol Based on Matured Ground Truth}
The above examination of mixed supervision and streaming adaptation suggests the need for a cleaner protocol. In the mixed-supervision setting, the use of POGT naturally induces a comparison between using revealed values as supervision and using the same values as additional input context for later direct prediction. In the streaming setting, supervision based on revealed ground truth is not necessarily matured relative to subsequent rolling predictions. These observations motivate a cleaner protocol that restricts adaptation to matured ground truth only.

The cleaner protocol is defined by the temporal status of the supervision it permits, rather than by a unique implementation rule. For example, one may update based on any subset of matured ground truth. Formally, denote the index set of matured mini-batches available at step $k$ as
\begin{equation}
\mathcal{M}_k := \{m < k : t_m + B_m + H - 1 \le t_k \}.
\end{equation}
Then a general adaptation objective based on matured ground truth may be written as
\begin{equation}
\mathcal{L}_k^{\mathrm{clean}} = \sum_{m \in \mathcal{S}_k} \omega_{k,m}\,\mathcal{L}_m^{\mathrm{matured}}, \qquad \mathcal{S}_k \subseteq \mathcal{M}_k,
\label{eq:clean_objective}
\end{equation}
where \(\mathcal{S}_k\) denotes the subset of matured mini-batches selected for adaptation at step \(k\), \(\omega_{k,m}\ge 0\) are combination weights, and
\begin{equation}
\mathcal{L}_m^{\mathrm{matured}} = \frac{1}{B_m} \sum_{j=1}^{B_m} \mathrm{MSE}(\hat{\mathbf{Y}}_m^{[j]}, \mathbf{Y}_m^{[j]}).
\end{equation}

Under our proposed protocol based solely on matured ground truth, predictions may still be organized in mini-batches, but the role of the mini-batch becomes different. Conceptually, the protocol no longer depends on supervision drawn from the current mini-batch; instead, adaptation is triggered only by previously matured targets. In this sense, the mini-batch mainly serves as an implementation-level grouping for prediction and evaluation. Moreover, although predictions within a mini-batch still overlap due to the rolling construction, the comparison between earlier and later samples in the overlapped region is no longer central to the protocol itself, since adaptation for each sample no longer relies on POGT from the current mini-batch. Rather, adaptation can already be triggered using matured past supervision before later samples in the current mini-batch are generated.

\section{Frequency-Aware Calibration}
With the cleaner protocol specifying which supervision signals should be used for adaptation, we now turn to the design of the adaptation module. Existing TSF-TTA adapters differ in their parameterization: TAFAS \citep{HyunGi2025} calibrates predictions through temporal-domain transformations, while PETSA \citep{medeiros2025accurate} recognizes the importance of spectral structure by incorporating a frequency-domain loss into a low-rank temporal calibration design. However, empirically, the resulting corrections in the frequency domain are often relatively smooth and less selective, even for PETSA with its frequency-domain loss. This motivates our Frequency-Aware Calibration (FAC), which directly parameterizes prediction correction in the frequency domain rather than only regularizing it through a spectral loss.

\begin{figure}[!htbp]
    \centering
    \includegraphics[width=0.96\linewidth]{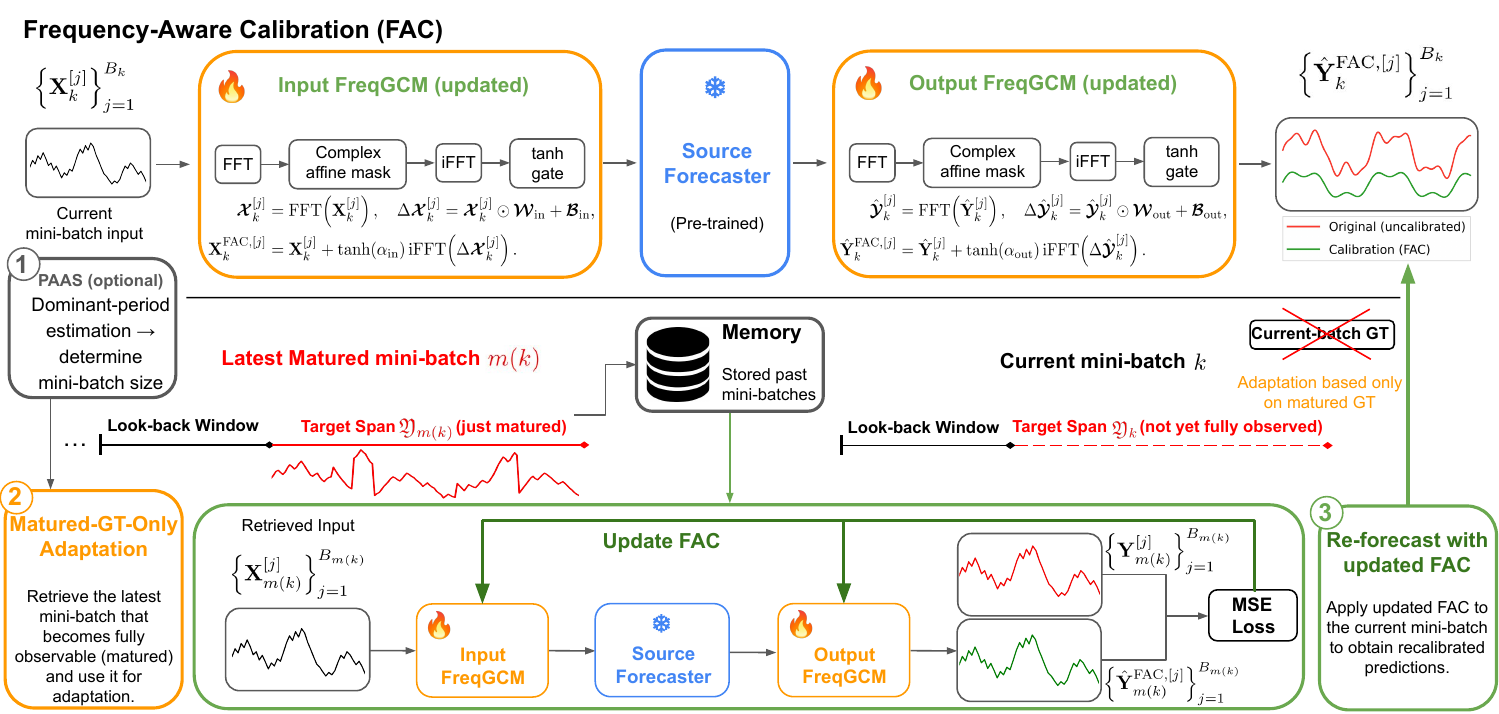}
    \caption{Overview of Frequency-Aware Calibration (FAC) under the proposed protocol. FAC uses input and output frequency-domain gated calibration modules (FreqGCMs), each applying a complex affine mask in the frequency domain followed by inverse FFT and gated residual correction. For simplicity, our implementation retrieves the most recent matured mini-batch for each update, after which the updated adapter is applied to re-forecast the current mini-batch. PAAS can optionally be used to determine the mini-batch size.}
    \label{fig:FAC_method}
\end{figure}

FAC adopts a minimalist design, since a central requirement for test-time adaptation in TSF is to keep the adapter lightweight. Figure~\ref{fig:FAC_method} provides an overview. Similar to TAFAS \citep{HyunGi2025} and PETSA \citep{medeiros2025accurate}, FAC places calibration modules before and after the frozen source forecaster. Each module performs a gated residual correction to the signal being calibrated, rather than directly replacing it. However, while TAFAS \citep{HyunGi2025} performs temporal-domain calibration through gated transformations and PETSA \citep{medeiros2025accurate} introduces spectral information mainly through an additional frequency-domain loss, our design directly parameterizes the calibration operation in the frequency domain via frequency-domain gated calibration modules (FreqGCMs). Specifically, FAC consists of an input FreqGCM and an output FreqGCM. Each FreqGCM first transforms the signal being calibrated into the frequency domain, applies a learnable complex affine mask, maps the frequency-domain correction back to the temporal domain through inverse FFT, and then controls the correction magnitude through a hyperbolic-tangent gate. On the input side, for the $j$-th sample in the $k$-th mini-batch, the input FreqGCM is formulated as
\begin{equation}
\begin{aligned}
\bm{\mathcal{X}}_{k}^{[j]}
&= \mathrm{FFT}\!\left(\mathbf{X}_k^{[j]}\right), \\
\Delta \bm{\mathcal{X}}_{k}^{[j]}
&=
\bm{\mathcal{X}}_{k}^{[j]} \odot \bm{\mathcal{W}}_{\mathrm{in}}
+
\bm{\mathcal{B}}_{\mathrm{in}}, \\
\mathbf{X}_{k}^{\mathrm{FAC},[j]}
&=
\mathbf{X}_k^{[j]}
+
\tanh(\alpha_{\mathrm{in}})
\,
\mathrm{iFFT}\!\left(
\Delta \bm{\mathcal{X}}_{k}^{[j]}
\right).
\end{aligned}
\end{equation}

The calibrated input is then passed to the frozen source forecaster,
\begin{equation}
\hat{\mathbf{Y}}_{k}^{[j]}
=
\mathcal{F}\!\left(
\mathbf{X}_{k}^{\mathrm{FAC},[j]}
\right).
\end{equation}
Similarly, the output FreqGCM calibrates the forecast in the frequency domain:
\begin{equation}
\begin{aligned}
\hat{\bm{\mathcal{Y}}}_{k}^{[j]}
&=
\mathrm{FFT}\!\left(
\hat{\mathbf{Y}}_k^{[j]}
\right), \\
\Delta \hat{\bm{\mathcal{Y}}}_{k}^{[j]}
&=
\hat{\bm{\mathcal{Y}}}_{k}^{[j]}
\odot
\bm{\mathcal{W}}_{\mathrm{out}}
+
\bm{\mathcal{B}}_{\mathrm{out}}, \\
\hat{\mathbf{Y}}_{k}^{\mathrm{FAC},[j]}
&=
\hat{\mathbf{Y}}_k^{[j]}
+
\tanh(\alpha_{\mathrm{out}})
\,
\mathrm{iFFT}\!\left(
\Delta \hat{\bm{\mathcal{Y}}}_{k}^{[j]}
\right).
\end{aligned}
\end{equation}
Here, the pairs $(\bm{\mathcal{W}}_{\mathrm{in}}, \bm{\mathcal{B}}_{\mathrm{in}})$ and $(\bm{\mathcal{W}}_{\mathrm{out}}, \bm{\mathcal{B}}_{\mathrm{out}})$ define element-wise complex affine transformations in the frequency domain, while $\alpha_{\mathrm{in}}$ and $\alpha_{\mathrm{out}}$ are learnable gating parameters. For notational simplicity, we write FFT and iFFT in the equations; in implementation, these transforms are realized by rFFT and irFFT so that the calibrated input and forecast remain real-valued. We use an element-wise affine form instead of a dense frequency-domain affine layer to keep the adapter lightweight and to preserve the frequency-wise interpretation of the correction. The multiplicative term allows each Fourier coefficient to be rescaled and phase-shifted, while the additive shift term allows the adapter to correct weak or near-zero frequency components that would be difficult to affect through purely multiplicative scaling. The residual form preserves the original signal when the learned correction is small, and the $\tanh$ gate controls the overall correction strength.

Recall that the proposed protocol permits adaptation using any subset of matured ground truth. For simplicity, we describe FAC using the most recent matured mini-batch for adaptation. The adaptation objective is
\begin{equation}
\mathcal{L}^{\mathrm{FAC}}_{m(k)}
=
\mathrm{MSE}\!\left(
\{\hat{\mathbf{Y}}^{\mathrm{FAC},[j]}_{m(k)}\}_{j=1}^{B_{m(k)}},
\{\mathbf{Y}^{[j]}_{m(k)}\}_{j=1}^{B_{m(k)}}
\right).
\end{equation}
Although additional regularization terms could be added to this objective, we keep the adaptation loss as a simple MSE on matured ground truth. This isolates the effect of frequency-domain parameterization and avoids conflating it with additional regularization choices. After the FAC modules are updated, the current mini-batch is re-forecast with the updated input and output FreqGCMs to obtain recalibrated predictions.

\section{Experiments}
\subsection{Experimental Settings}
We evaluate the proposed FAC on six multivariate TSF benchmark datasets: ETTh1, ETTh2, ETTm1, ETTm2, Weather, and Exchange. For all datasets, we use a chronological train/validation/test split with ratio $(0.7, 0.1, 0.2)$. The look-back length is set to $L=96$, and the prediction horizon is chosen from $H \in \{96, 192, 336, 720\}$, except for Exchange, where we use $H \in \{96, 192, 336\}$ because the full target horizon for $H=720$ does not mature before the end of the test sequence. Consistent with existing TSF-TTA studies, we evaluate five representative source forecasters: iTransformer \citep{liu2024itransformer}, PatchTST \citep{nie2023a}, DLinear \citep{Zeng2023DLinear}, OLS \citep{toner2024an}, and FreTS \citep{yi2023frequencydomain} \footnote{We omit MICN \citep{wang2023micn} from the main experiments because its architecture categorization is ambiguous in prior TSF-TTA comparisons. Although some studies group it with MLP-based forecasters, MICN is more accurately viewed as a multi-scale convolution-based model.}.

We compare FAC with the frozen source forecaster without TTA, TAFAS \citep{HyunGi2025}, and PETSA \citep{medeiros2025accurate} under the proposed protocol based solely on matured ground truth. For fairness and simplicity, only the most recent matured mini-batch is used for adaptation across methods. Since the current mini-batch does not provide supervision under this protocol, the POGT-based partial prediction adjustment in the original TAFAS protocol is no longer used. Instead, after the adapter is updated on the most recent matured mini-batch, the current mini-batch is re-forecast and the full prediction horizon is evaluated using the updated adapter. In particular, for the $j$-th sample in the $k$-th mini-batch, we use
\begin{equation}
\tilde{\mathbf{Y}}^{[j]}_k
=
\hat{\mathbf{Y}}^{\mathrm{adp},[j]}_k,
\quad j=1,\ldots,B_k,
\end{equation}
rather than a partial stitching rule between pre-update and post-update predictions \footnote{We do not include a matured-only COSA \citep{im2026cosa} variant in the main experiments because COSA is designed around an output-space streaming adapter with context constructed from revealed target statistics. Adapting it to our proposed protocol would require nontrivial design choices.}. Hyperparameters are taken from the officially released code whenever available. All experiments were conducted on an NVIDIA GeForce RTX 4070 Super 12GB and an NVIDIA GeForce RTX 4090 24GB.

\subsection{Main Results}
\subsubsection{Forecasting Performance}
Table~\ref{tab:main_results} reports the MSE results across six datasets, five source forecasters, and four prediction horizons. All TSF-TTA methods are evaluated under the proposed protocol using only matured ground truth. 

\begin{table}[!htbp]
\centering
\caption{Forecasting performance under the proposed protocol using only matured ground truth. Lower MSE is better. The best result within each backbone and horizon is highlighted in red bold, and the second-best result is underlined.}
\label{tab:main_results}
\small
\setlength{\tabcolsep}{1.5pt}
\renewcommand{\arraystretch}{1.5}
\resizebox{\textwidth}{!}{
\begin{tabular}{ll|cccc|cccc|cccc|cccc|cccc}
\hline
\textbf{Dataset} & $H$
& \multicolumn{4}{c|}{\textbf{DLinear}}
& \multicolumn{4}{c|}{\textbf{FreTS}}
& \multicolumn{4}{c|}{\textbf{iTransformer}}
& \multicolumn{4}{c|}{\textbf{OLS}}
& \multicolumn{4}{c}{\textbf{PatchTST}} \\
\cline{3-22}
&
& Base & TAFAS & PETSA & \textbf{FAC}
& Base & TAFAS & PETSA & \textbf{FAC}
& Base & TAFAS & PETSA & \textbf{FAC}
& Base & TAFAS & PETSA & \textbf{FAC}
& Base & TAFAS$^\dagger$ & PETSA & \textbf{FAC} \\
\hline
\textbf{ETTh1} & 96
& 0.4695 & 0.4617 & \second{0.4613} & \best{0.4554}
& 0.4461 & 0.4391 & \second{0.4383} & \best{0.4357}
& 0.4482 & 0.4421 & \best{0.4394} & \second{0.4400}
& 0.4510 & \second{0.4428} & 0.4449 & \best{0.4382}
& 0.4324 & 0.4270 & \second{0.4255} & \best{0.4241} \\
& 192
& 0.5213 & \second{0.5112} & 0.5123 & \best{0.5089}
& 0.5022 & \second{0.4943} & 0.4960 & \best{0.4934}
& 0.5091 & 0.5020 & \second{0.5014} & \best{0.5010}
& 0.5045 & \second{0.4929} & 0.5007 & \best{0.4921}
& 0.4900 & \second{0.4818} & \second{0.4818} & \best{0.4797} \\
& 336
& 0.5659 & \second{0.5595} & 0.5626 & \best{0.5582}
& 0.5544 & \best{0.5483} & 0.5530 & \second{0.5498}
& 0.5662 & \second{0.5624} & 0.5641 & \best{0.5587}
& 0.5510 & \second{0.5427} & 0.5488 & \best{0.5422}
& 0.5608 & \best{0.5506} & 0.5570 & \second{0.5525} \\
& 720
& 0.7117 & \second{0.6926} & 0.6938 & \best{0.6891}
& 0.7181 & \best{0.6857} & \second{0.6908} & 0.7067
& 0.7007 & \second{0.6755} & 0.6763 & \best{0.6687}
& 0.6996 & \best{0.6704} & 0.6805 & \second{0.6780}
& 0.7112 & \best{0.6807} & \second{0.7019} & 0.7096 \\
\hline
\textbf{ETTh2} & 96
& 0.2323 & \second{0.2308} & 0.2312 & \best{0.2288}
& 0.2385 & \second{0.2363} & 0.2366 & \best{0.2338}
& 0.2642 & 0.2620 & \second{0.2602} & \best{0.2561}
& 0.2306 & \second{0.2289} & 0.2293 & \best{0.2274}
& 0.2385 & \second{0.2380} & 0.2383 & \best{0.2374} \\
& 192
& 0.2862 & 0.2830 & \second{0.2820} & \best{0.2819}
& 0.2867 & \second{0.2830} & 0.2844 & \best{0.2814}
& 0.3081 & \second{0.3004} & 0.3005 & \best{0.2970}
& 0.2838 & \second{0.2821} & 0.2829 & \best{0.2796}
& 0.2840 & \second{0.2752} & 0.2810 & \best{0.2747} \\
& 336
& 0.3252 & \second{0.3175} & 0.3190 & \best{0.3144}
& 0.3318 & \second{0.3195} & 0.3250 & \best{0.3172}
& 0.3402 & \second{0.3290} & 0.3312 & \best{0.3284}
& 0.3257 & \second{0.3189} & 0.3240 & \best{0.3137}
& 0.3233 & \second{0.3108} & 0.3190 & \best{0.3079} \\
& 720
& 0.4087 & \best{0.3825} & \second{0.3884} & \best{0.3825}
& 0.4118 & \second{0.3814} & 0.3968 & \best{0.3788}
& 0.4261 & \best{0.3959} & 0.4049 & \second{0.3982}
& 0.4161 & \second{0.3859} & 0.3927 & \best{0.3854}
& 0.4292 & \second{0.4065} & 0.4203 & \best{0.4002} \\
\hline
\textbf{ETTm1} & 96
& 0.3715 & \second{0.3504} & 0.3541 & \best{0.3465}
& 0.3676 & \second{0.3559} & 0.3572 & \best{0.3517}
& 0.3823 & \second{0.3680} & 0.3700 & \best{0.3459}
& 0.3709 & \second{0.3542} & 0.3591 & \best{0.3462}
& 0.4041 & 0.3826 & \second{0.3817} & \best{0.3530} \\
& 192
& 0.4438 & \second{0.4165} & 0.4192 & \best{0.4139}
& 0.4325 & 0.4191 & \second{0.4189} & \best{0.4183}
& 0.4407 & \second{0.4189} & \second{0.4189} & \best{0.4087}
& 0.4437 & \second{0.4184} & 0.4252 & \best{0.4140}
& 0.4515 & \second{0.4370} & 0.4405 & \best{0.4213} \\
& 336
& 0.5183 & \second{0.4801} & 0.4831 & \best{0.4791}
& 0.5006 & \second{0.4810} & \best{0.4793} & 0.4836
& 0.5077 & 0.4800 & \second{0.4788} & \best{0.4781}
& 0.5179 & \second{0.4811} & 0.4896 & \best{0.4794}
& 0.5044 & \second{0.4864} & 0.4970 & \best{0.4787} \\
& 720
& 0.5929 & \second{0.5521} & 0.5600 & \best{0.5494}
& 0.5703 & \best{0.5472} & \second{0.5523} & 0.5533
& 0.6021 & \second{0.5579} & 0.5615 & \best{0.5460}
& 0.5923 & \second{0.5515} & 0.5594 & \best{0.5497}
& 0.5608 & \second{0.5413} & 0.5479 & \best{0.5399} \\
\hline
\textbf{ETTm2} & 96
& 0.1598 & \second{0.1581} & 0.1583 & \best{0.1560}
& 0.1582 & \second{0.1566} & 0.1569 & \best{0.1563}
& 0.1647 & \second{0.1640} & \second{0.1640} & \best{0.1613}
& 0.1602 & \second{0.1594} & 0.1595 & \best{0.1563}
& 0.1581 & 0.1569 & \second{0.1569} & \best{0.1558} \\
& 192
& 0.1930 & \second{0.1917} & 0.1920 & \best{0.1894}
& 0.1923 & \best{0.1901} & \second{0.1908} & \best{0.1901}
& 0.2189 & 0.2164 & \second{0.2157} & \best{0.2067}
& 0.1935 & \second{0.1924} & 0.1925 & \best{0.1897}
& 0.2097 & \second{0.2058} & 0.2079 & \best{0.1976} \\
& 336
& 0.2324 & \second{0.2299} & 0.2303 & \best{0.2286}
& 0.2320 & \best{0.2291} & 0.2312 & \second{0.2294}
& 0.2808 & 0.2735 & \second{0.2653} & \best{0.2623}
& 0.2331 & \second{0.2308} & 0.2316 & \best{0.2289}
& 0.2482 & 0.2482 & \second{0.2473} & \best{0.2422} \\
& 720
& 0.3062 & \second{0.2993} & \best{0.2984} & \best{0.2984}
& 0.3013 & \best{0.2924} & 0.2983 & \second{0.2955}
& 0.3457 & 0.3303 & \best{0.3263} & \second{0.3290}
& 0.3066 & \second{0.2996} & 0.3003 & \best{0.2973}
& 0.3283 & \second{0.3218} & 0.3240 & \best{0.3192} \\
\hline
\textbf{Weather} & 96
& 0.1954 & \second{0.1803} & 0.1812 & \best{0.1797}
& 0.1855 & \second{0.1757} & 0.1795 & \best{0.1678}
& 0.1724 & \second{0.1662} & 0.1662 & \best{0.1605}
& 0.1957 & \second{0.1825} & 0.1890 & \best{0.1728}
& 0.1746 & 0.1700 & \second{0.1676} & \best{0.1635} \\
& 192
& 0.2403 & \second{0.2265} & 0.2308 & \best{0.2231}
& 0.2309 & \second{0.2167} & 0.2229 & \best{0.2114}
& 0.2239 & \second{0.2129} & 0.2165 & \best{0.2103}
& 0.2406 & \second{0.2238} & 0.2293 & \best{0.2171}
& 0.2188 & 0.2119 & \second{0.2120} & \best{0.2050} \\
& 336
& 0.2918 & \second{0.2726} & 0.2776 & \best{0.2717}
& 0.2842 & \second{0.2661} & 0.2720 & \best{0.2621}
& 0.2806 & \second{0.2629} & 0.2672 & \best{0.2618}
& 0.2921 & \second{0.2729} & 0.2785 & \best{0.2665}
& 0.2771 & \second{0.2659} & 0.2703 & \best{0.2570} \\
& 720
& 0.3643 & \second{0.3509} & 0.3557 & \best{0.3426}
& 0.3599 & \second{0.3423} & 0.3550 & \best{0.3376}
& 0.3563 & \second{0.3418} & 0.3470 & \best{0.3357}
& 0.3645 & \second{0.3467} & 0.3598 & \best{0.3401}
& 0.3555 & \second{0.3356} & 0.3529 & \best{0.3341} \\
\hline
\textbf{Exchange} & 96
& 0.0913 & \best{0.0875} & \second{0.0898} & \best{0.0875}
& 0.0828 & \best{0.0796} & 0.0825 & \second{0.0810}
& 0.0882 & \second{0.0873} & 0.0881 & \best{0.0867}
& 0.0814 & \second{0.0799} & 0.0813 & \best{0.0793}
& 0.0854 & \second{0.0816} & 0.0850 & \best{0.0813} \\
& 192
& 0.1827 & \best{0.1732} & 0.1787 & \second{0.1758}
& 0.1734 & \best{0.1654} & 0.1729 & \second{0.1667}
& 0.1841 & \second{0.1800} & 0.1820 & \best{0.1773}
& 0.1727 & \second{0.1650} & 0.1719 & \best{0.1648}
& 0.1774 & \best{0.1666} & 0.1765 & \second{0.1692} \\
& 336
& 0.3277 & \second{0.3052} & 0.3154 & \best{0.2985}
& 0.3241 & \second{0.3180} & 0.3223 & \best{0.3166}
& 0.3157 & \second{0.3014} & 0.3105 & \best{0.2983}
& 0.3225 & \best{0.2846} & 0.3058 & \second{0.2924}
& 0.3382 & \second{0.3048} & 0.3188 & \best{0.3011} \\
\hline
\end{tabular}
}
\vspace{0.5em}

\begin{minipage}{0.98\textwidth}
\footnotesize
\noindent\textit{Note.}
FAC denotes our proposed method. For PatchTST, TAFAS$^\dagger$ follows the original PatchTST-specific implementation with single-channel calibration and no input calibration. PETSA uses multivariate calibration on PatchTST to match its frequency-domain and distributional adaptation objectives over multivariate forecasts.
\end{minipage}

\end{table}

Overall, FAC achieves the best or second-best performance in most settings and often outperforms TAFAS and PETSA under the same protocol using only matured ground truth. In addition, all three adaptation methods generally improve over the frozen source forecaster, suggesting that matured ground truth can serve as an effective adaptation signal for TSF-TTA.

\subsubsection{Parameter Efficiency}
Besides forecasting error, parameter efficiency is a central consideration when designing a test-time adapter. A direct way to measure parameter efficiency is to compare the number of trainable parameters in the calibration modules. Table~\ref{tab:param_efficiency} reports the trainable parameter counts of different adapters under our experimental settings. FAC consistently requires the fewest trainable parameters across datasets and horizons.

\begin{table}[!htbp]
\centering
\caption{Trainable parameter counts of different test-time adapters. For TAFAS, we report both the standard count and the PatchTST-specific count, since TAFAS uses channel dimension $1$ for PatchTST but the number of variables $C$ for other backbones. The smallest count in each row is highlighted in red bold, and the second smallest is underlined.}
\label{tab:param_efficiency}
\small
\setlength{\tabcolsep}{6pt}
\renewcommand{\arraystretch}{1.25}
\begin{tabular}{cc|cccc}
\hline
\rule[-5pt]{0pt}{21pt}
\raisebox{4pt}[0pt][0pt]{\textbf{Dataset}}
& \raisebox{4pt}[0pt][0pt]{\textbf{$H$}}
& \shortstack{\textbf{TAFAS}\\[-1pt]\vphantom{\textbf{(PatchTST)}}}
& \shortstack{\textbf{TAFAS}\\[-1pt]\textbf{(PatchTST)}}
& \shortstack{\textbf{PETSA}\\[-1pt]\vphantom{\textbf{(PatchTST)}}}
& \shortstack{\textbf{FAC}\\[-1pt]\textbf{(Ours)}} \\
\hline
\raisebox{-9pt}[0pt][0pt]{\shortstack{\textbf{ETT}\\[-1pt]\textbf{(h1/2, m1/2)}}}
& 96  & 130382  & \second{18626}  & 25934  & \best{2758} \\
& 192 & 324590  & 46370  & \second{38894}  & \best{4102} \\
& 336 & 857822  & 122546 & \second{58334}  & \best{6118} \\
& 720 & 3699038 & 528434 & \second{110174} & \best{11494} \\
\hline
\textbf{Weather} & 96  & 391146   & \second{18626}  & 71658  & \best{8274} \\
                 & 192 & 973770   & \second{46370}  & 107466 & \best{12306} \\
                 & 336 & 2573466  & \second{122546} & 161178 & \best{18354} \\
                 & 720 & 11097114 & 528434 & \second{304410} & \best{34482} \\
\hline
\textbf{Exchange} & 96  & 149008  & \second{18626}  & 29200  & \best{3152} \\
                  & 192 & 370960  & 46370  & \second{43792}  & \best{4688} \\
                  & 336 & 980368  & 122546 & \second{65680}  & \best{6992} \\
\hline
\end{tabular}
\end{table}

Table~\ref{tab:param_order} summarizes how the adapter size scales with the look-back length $L$, prediction horizon $H$, number of variables $C$, and PETSA rank $r$. TAFAS scales quadratically with the sequence lengths due to its dense temporal calibration, whereas PETSA achieves rank-dependent linear scaling through low-rank adapters. FAC instead uses an element-wise frequency-domain affine parameterization, leading to linear scaling in $C$, $L$, and $H$ without the rank factor. Among the compared adapters, FAC has the most favorable scaling behavior.

\begin{table}[!htbp]
\centering
\caption{Parameter scaling of different test-time adapters.}
\label{tab:param_order}
\small
\setlength{\tabcolsep}{8pt}
\renewcommand{\arraystretch}{1.2}
\begin{tabular}{c|c}
\hline
\textbf{Adapter} & \textbf{Parameter scaling} \\ 
\hline
\textbf{TAFAS} & $O\!\left(C(L^2 + H^2)\right)$ \\
\textbf{TAFAS (PatchTST)} & $O\!\left(L^2 + H^2\right)$ \\
\textbf{PETSA} & $O\!\left(Cr(L+H)\right)$ \\
\textbf{FAC (Ours)}   & $O\!\left(C(L+H)\right)$ \\
\hline
\end{tabular}
\end{table}

\subsection{Analysis}
\subsubsection{Is POGT Necessary for Effective Adaptation?}
Beyond the protocol-level motivation discussed above, we further examine whether POGT provides a consistent empirical benefit for adaptation. To this end, we compare the original TAFAS protocol, which uses mixed supervision, with a variant that uses only matured ground truth. As shown in Table~\ref{tab:tafas_full_only_etth1}, the original TAFAS protocol often achieves slightly lower MSE, but the variant using matured ground truth remains close across source forecasters and prediction horizons. These results suggest that POGT is not consistently necessary for effective adaptation. Together with the protocol-level motivation, this supports using the matured-GT variant as a cleaner protocol for our main experiments. Complete results for TAFAS are provided in Appendix~\ref{app:full_only_tafas}.

\begin{table}[!htbp]
\centering
\caption{Original TAFAS vs. matured-only TAFAS on ETTh1. Lower MSE is better. The better result in each pair is highlighted in red bold.}
\label{tab:tafas_full_only_etth1}
\small
\setlength{\tabcolsep}{4pt}
\renewcommand{\arraystretch}{1.15}
\begin{tabular}{c|cc|cc|cc|cc|cc}
\hline
$H$
& \multicolumn{2}{c|}{DLinear}
& \multicolumn{2}{c|}{FreTS}
& \multicolumn{2}{c|}{iTransformer}
& \multicolumn{2}{c|}{OLS}
& \multicolumn{2}{c}{PatchTST} \\
\cline{2-11}
& Orig. & Matured
& Orig. & Matured
& Orig. & Matured
& Orig. & Matured
& Orig. & Matured \\
\hline
96  
& 0.4618 & \best{0.4617}
& 0.4394 & \best{0.4391}
& \best{0.4398} & 0.4421
& \best{0.4419} & 0.4428
& \best{0.4261} & 0.4270 \\
192 
& 0.5117 & \best{0.5112}
& \best{0.4940} & 0.4943
& \best{0.5017} & 0.5020
& \best{0.4924} & 0.4929
& \best{0.4809} & 0.4818 \\
336 
& 0.5604 & \best{0.5595}
& \best{0.5475} & 0.5483
& 0.5643 & \best{0.5624}
& \best{0.5415} & 0.5427
& \best{0.5501} & 0.5506 \\
720 
& \best{0.6820} & 0.6926
& 0.6872 & \best{0.6857}
& \best{0.6591} & 0.6755
& \best{0.6654} & 0.6704
& 0.6815 & \best{0.6807} \\
\hline
\end{tabular}
\end{table}
\subsubsection{Frequency-Domain Behavior of Prediction Corrections}

We further examine how different adapters modify the forecast in the frequency domain. 
This analysis is not intended to measure forecasting error directly; instead, it characterizes the frequency-domain structure of the corrections induced by different adaptation methods. 
Specifically, we compute the difference between predictions before and after adaptation, apply rFFT along the forecasting horizon, and average the correction magnitude over samples and channels.

\begin{figure}[!htbp]
    \centering    
    \includegraphics[width=\linewidth]{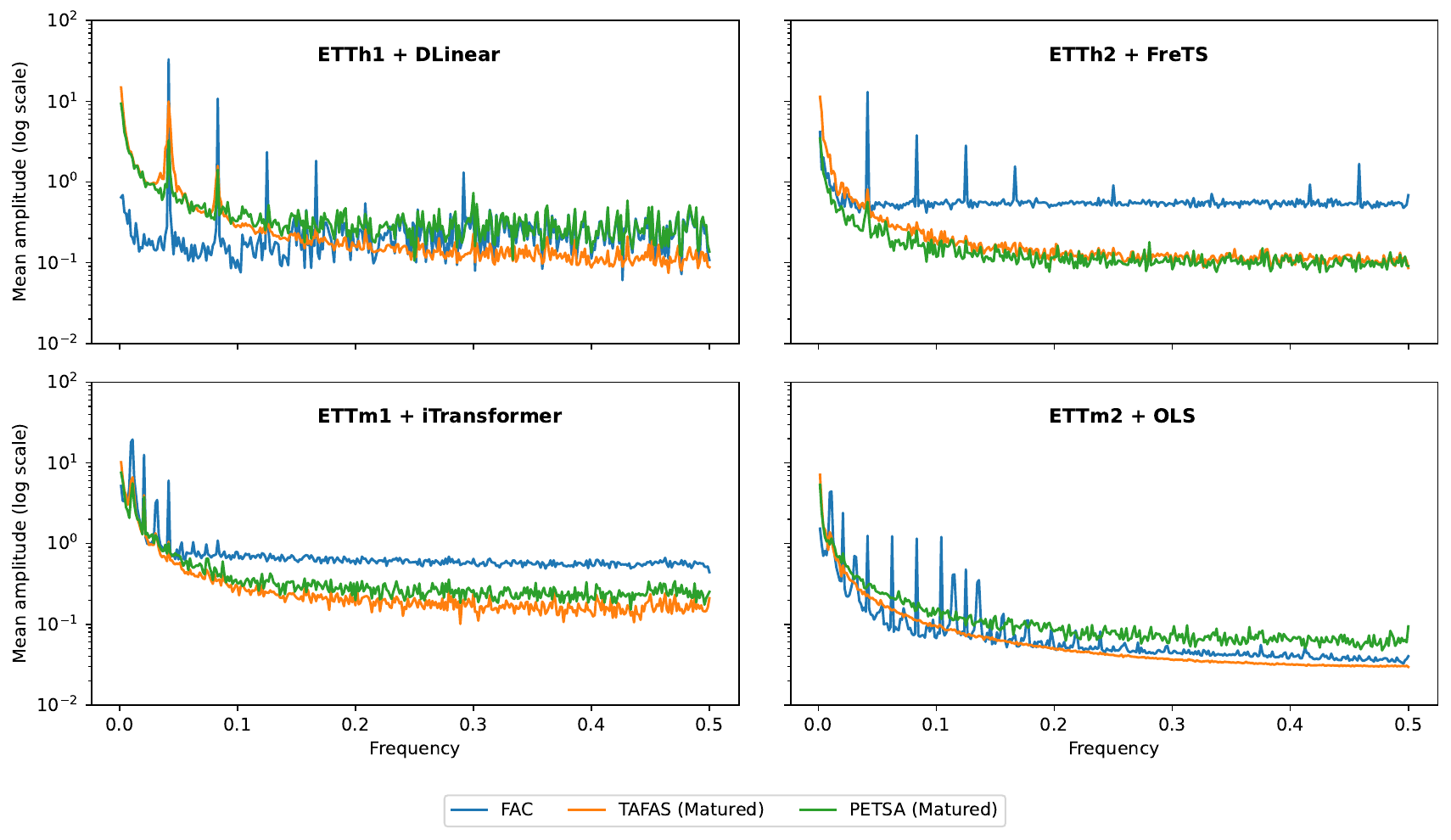}
    \caption{Frequency-domain magnitude of prediction corrections under the matured-only protocol for forecasting horizon $H=720$. The plotted spectra are computed from the differences between predictions after and before adaptation. The x-axis denotes frequency, while the y-axis shows the correction magnitude on a logarithmic scale. The figure characterizes adaptation-induced corrections rather than forecasting error; a larger magnitude does not necessarily indicate a more accurate forecast.}
    \label{fig:FAC_freq}
\end{figure}

Figure~\ref{fig:FAC_freq} illustrates the resulting correction spectra for representative dataset--forecaster pairs at forecasting horizon $H=720$, with more comprehensive results across horizons deferred to Appendix~\ref{app:freq_spectra}. 
As shown in the figure, TAFAS and PETSA often produce relatively smooth frequency-domain corrections, with magnitudes changing gradually across frequencies. 
In contrast, FAC exhibits more localized peaks, suggesting that its corrections are more selective with respect to specific frequency components. 
This behavior is consistent with the motivation of FAC: by directly applying an element-wise complex affine mask in the frequency domain, FAC allows individual Fourier components to be adjusted separately, which can naturally lead to more localized spectral corrections. Therefore, the observed peaks should be interpreted as evidence of frequency-selective adaptation rather than as a direct measure of forecasting accuracy; a larger correction magnitude does not necessarily imply a more accurate forecast.

\section{Conclusions, Limitations, and Future Work}

\textbf{Conclusions.}
We re-examine the adaptation protocols used in existing TSF-TTA methods. Specifically, we revisit the mixed-supervision protocol that incorporates both POGT and matured ground truth, as well as the streaming adaptation protocol based on revealed targets, and discuss their potential protocol-level limitations. To improve protocol cleanliness, we propose a more principled adaptation protocol that uses only matured ground truth. 
We further diagnose existing adapters in the frequency domain and propose FAC, a lightweight adapter that directly parameterizes prediction corrections in the frequency domain. Experimental results show that FAC achieves competitive and consistent forecasting performance across diverse settings while requiring substantially fewer trainable parameters.

\textbf{Limitations.}
Although the proposed protocol improves protocol cleanliness by using only matured ground truth, it uses a relatively strict definition of maturation at the mini-batch level. The adapter is updated only after the full target span of a past mini-batch matures, which may introduce additional feedback delay when the forecasting horizon or mini-batch size is large. 

Although FAC requires significantly fewer trainable parameters, this parameter efficiency does not always translate into lower wall-clock runtime.
As shown in Appendix~\ref{app:runtime}, under our current implementation that adapts using only the most recent matured mini-batch, rFFT/iFFT operations and additional tensor transformations can introduce overhead compared with TAFAS in some settings.

\textbf{Future Work.}
Future work could explore more flexible matured-supervision implementations that do not require waiting for the entire mini-batch to mature before using already observable targets. Another direction is to use multiple available matured mini-batches, as allowed by Eq.~\ref{eq:clean_objective}, instead of only the most recent one. When multiple matured mini-batches are used, cached frequency-domain representations of historical matured mini-batches may help reduce repeated FFT computations and improve the runtime efficiency of frequency-aware adaptation. Together, these directions could make adaptation based on matured ground truth more proactive and efficient while preserving the cleanliness of the proposed protocol.

\bibliography{main}
\bibliographystyle{tmlr}

\clearpage
\appendix
\renewcommand{\thetable}{\thesection.\arabic{table}}
\renewcommand{\thefigure}{\thesection.\arabic{figure}}
\setcounter{table}{0}
\setcounter{figure}{0}

\section{Additional Experimental Results}
\label{app:additional_results}

\subsection{Additional Early-versus-Late Comparison}
\label{app:early_late_additional}

\begin{figure*}[!htbp]
    \centering
    \includegraphics[width=0.95\textwidth]{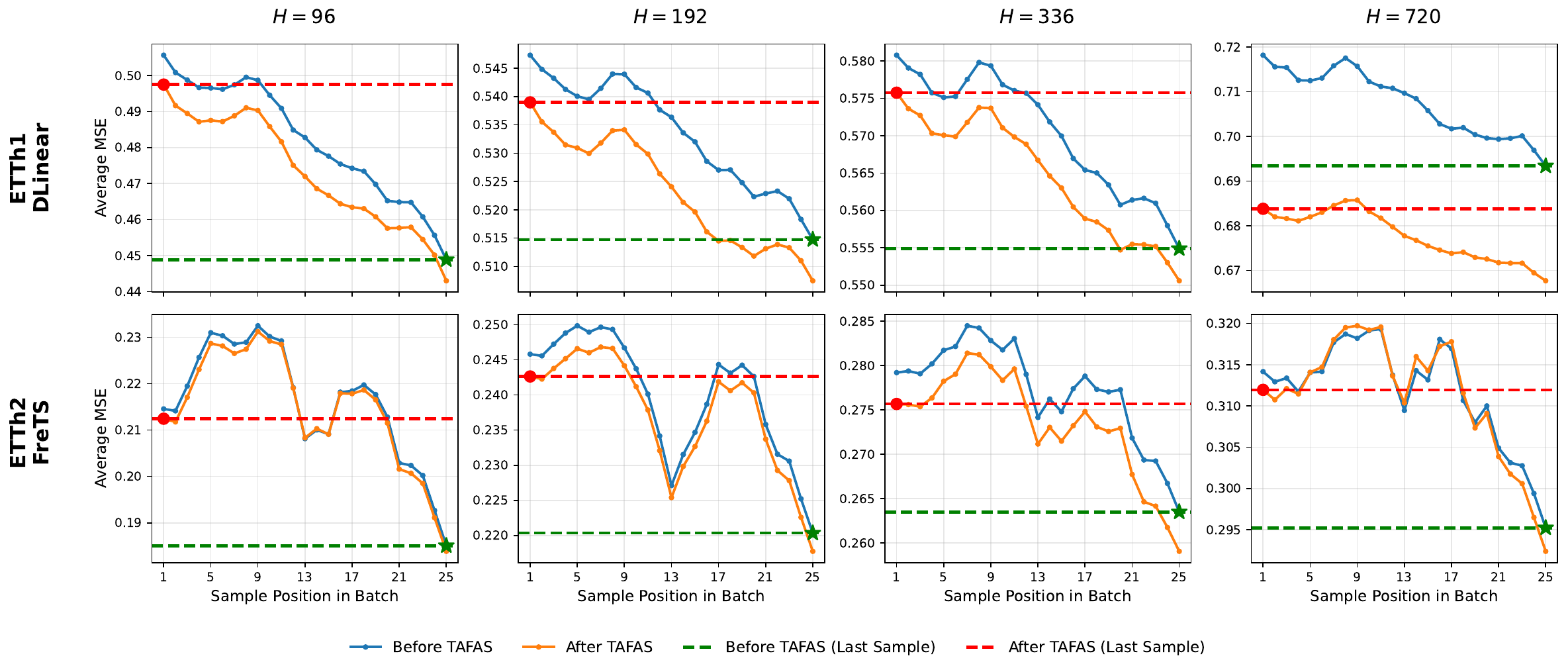}
    \caption{
    Additional early-versus-late comparison with mini-batch size $B_k=25$ on ETTh1 with DLinear and ETTh2 with FreTS.
    The curves show average MSE across sample positions within a mini-batch. The dashed horizontal lines mark the before-adaptation and after-adaptation MSE of the last sample for visual reference.
    }
    \label{fig:early_late_b25}
\end{figure*}

Figure~\ref{fig:early_late_b25} provides an additional early-versus-late comparison for $B_k=25$, which is the second most common mini-batch size induced by PAAS in our experiments after $B_k=97$. Compared with the $B_k=97$ setting, this smaller mini-batch size gives later samples less additional revealed context, making the comparison relatively more favorable to earlier adjusted predictions. Nevertheless, later direct predictions still generally achieve comparable or lower MSE than the adjusted predictions of earlier samples. This further supports our argument that using revealed values as POGT supervision is not consistently more beneficial than using the same values as input context in later rolling windows.

\subsection{Effect of Using Only Matured Ground Truth}
\label{app:full_only_tafas}

Table~\ref{tab:orig_vs_matured_tafas} provides the full comparison between the original TAFAS protocol and its variant using only matured ground truth. The original TAFAS protocol, which incorporates POGT through mixed supervision, often achieves slightly lower MSE. However, this advantage is not consistent across datasets, horizons, and source forecasters, and the matured-only variant remains close in most settings. These results complement the analysis in the main text and suggest that removing POGT-based supervision does not lead to substantial performance degradation, while yielding a cleaner adaptation protocol.

\begingroup
\scriptsize
\setlength{\tabcolsep}{4pt}
\renewcommand{\arraystretch}{1.35}

\begin{longtable}{cc|cc|cc|cc|cc|cc}
\caption{
Original TAFAS versus its variant using only matured ground truth across datasets and source forecasters.
Lower MSE is highlighted within each original/matured pair.
}
\label{tab:orig_vs_matured_tafas}\\

\hline
Dataset & Horizon
& \multicolumn{2}{c|}{DLinear}
& \multicolumn{2}{c|}{FreTS}
& \multicolumn{2}{c|}{iTransformer}
& \multicolumn{2}{c|}{OLS}
& \multicolumn{2}{c}{PatchTST} \\
\cline{3-12}
&
& Orig. & Matured
& Orig. & Matured
& Orig. & Matured
& Orig. & Matured
& Orig. & Matured \\
\hline
\endfirsthead

\hline
Dataset & Horizon
& \multicolumn{2}{c|}{DLinear}
& \multicolumn{2}{c|}{FreTS}
& \multicolumn{2}{c|}{iTransformer}
& \multicolumn{2}{c|}{OLS}
& \multicolumn{2}{c}{PatchTST} \\
\cline{3-12}
&
& Orig. & Matured
& Orig. & Matured
& Orig. & Matured
& Orig. & Matured
& Orig. & Matured \\
\hline
\endhead

\hline
\endfoot

ETTh1 & 96
& 0.4618 & \best{0.4617}
& 0.4394 & \best{0.4391}
& \best{0.4398} & 0.4421
& \best{0.4419} & 0.4428
& \best{0.4261} & 0.4270 \\

& 192
& 0.5117 & \best{0.5112}
& \best{0.4940} & 0.4943
& \best{0.5017} & 0.5020
& \best{0.4924} & 0.4929
& \best{0.4809} & 0.4818 \\

& 336
& 0.5604 & \best{0.5595}
& \best{0.5475} & 0.5483
& 0.5643 & \best{0.5624}
& \best{0.5415} & 0.5427
& \best{0.5501} & 0.5506 \\

& 720
& \best{0.6820} & 0.6926
& 0.6872 & \best{0.6857}
& \best{0.6591} & 0.6755
& \best{0.6654} & 0.6704
& 0.6815 & \best{0.6807} \\
\hline

ETTh2 & 96
& \best{0.2303} & 0.2308
& \best{0.2362} & 0.2363
& \best{0.2613} & 0.2620
& \best{0.2284} & 0.2289
& \best{0.2378} & 0.2380 \\

& 192
& 0.2842 & \best{0.2830}
& \best{0.2828} & 0.2830
& \best{0.2984} & 0.3004
& \best{0.2817} & 0.2821
& \best{0.2743} & 0.2752 \\

& 336
& 0.3185 & \best{0.3175}
& 0.3204 & \best{0.3195}
& \best{0.3286} & 0.3290
& \best{0.3186} & 0.3189
& 0.3119 & \best{0.3108} \\

& 720
& 0.3873 & \best{0.3825}
& 0.3831 & \best{0.3814}
& 0.3981 & \best{0.3959}
& 0.3906 & \best{0.3859}
& \best{0.4023} & 0.4065 \\
\hline
\pagebreak[4]

ETTm1 & 96
& \best{0.3497} & 0.3504
& 0.3574 & \best{0.3559}
& \best{0.3658} & 0.3680
& \best{0.3521} & 0.3542
& 0.3912 & \best{0.3826} \\

& 192
& 0.4166 & \best{0.4165}
& 0.4205 & \best{0.4191}
& \best{0.4182} & 0.4189
& \best{0.4164} & 0.4184
& \best{0.4366} & 0.4370 \\

& 336
& \best{0.4799} & 0.4801
& 0.4816 & \best{0.4810}
& \best{0.4780} & 0.4800
& \best{0.4786} & 0.4811
& \best{0.4863} & 0.4864 \\

& 720
& \best{0.5488} & 0.5521
& 0.5474 & \best{0.5472}
& \best{0.5529} & 0.5579
& \best{0.5479} & 0.5515
& 0.5414 & \best{0.5413} \\
\hline

ETTm2 & 96
& 0.1584 & \best{0.1581}
& 0.1571 & \best{0.1566}
& \best{0.1639} & 0.1640
& \best{0.1592} & 0.1594
& 0.1575 & \best{0.1569} \\

& 192
& \best{0.1913} & 0.1917
& 0.1907 & \best{0.1901}
& \best{0.2159} & 0.2164
& \best{0.1921} & 0.1924
& 0.2063 & \best{0.2058} \\

& 336
& \best{0.2289} & 0.2299
& 0.2292 & \best{0.2291}
& 0.2751 & \best{0.2735}
& \best{0.2299} & 0.2308
& 0.2487 & \best{0.2482} \\

& 720
& \best{0.2968} & 0.2993
& \best{0.2918} & 0.2924
& \best{0.3278} & 0.3303
& \best{0.2982} & 0.2996
& 0.3223 & \best{0.3218} \\
\hline

Weather & 96
& \best{0.1796} & 0.1803
& \best{0.1744} & 0.1757
& \best{0.1647} & 0.1662
& \best{0.1810} & 0.1825
& 0.1715 & \best{0.1700} \\

& 192
& \best{0.2244} & 0.2265
& \best{0.2151} & 0.2167
& \best{0.2119} & 0.2129
& \best{0.2221} & 0.2238
& 0.2130 & \best{0.2119} \\

& 336
& \best{0.2709} & 0.2726
& \best{0.2640} & 0.2661
& \best{0.2615} & 0.2629
& \best{0.2707} & 0.2729
& 0.2660 & \best{0.2659} \\

& 720
& \best{0.3500} & 0.3509
& \best{0.3400} & 0.3423
& \best{0.3410} & 0.3418
& \best{0.3443} & 0.3467
& 0.3365 & \best{0.3356} \\
\hline

Exchange & 96
& 0.0885 & \best{0.0875}
& \best{0.0790} & 0.0796
& \best{0.0867} & 0.0873
& \best{0.0795} & 0.0799
& \best{0.0813} & 0.0816 \\

& 192
& 0.1760 & \best{0.1732}
& \best{0.1639} & 0.1654
& \best{0.1760} & 0.1800
& \best{0.1643} & 0.1650
& 0.1673 & \best{0.1666} \\

& 336
& \best{0.2941} & 0.3052
& \best{0.2945} & 0.3180
& \best{0.2937} & 0.3014
& \best{0.2835} & 0.2846
& \best{0.2807} & 0.3048 \\
\hline

\end{longtable}
\endgroup

\subsection{Is Output-Only Calibration Sufficient?}
\label{app:input_output_ablation}

We compare the standard input-output calibration with an output-only variant across datasets, horizons, and source forecasters. The goal is to examine whether calibrating the look-back input window provides additional benefit beyond calibrating only the forecast output. Lower MSE is highlighted within each input-output versus output-only pair.

As shown in Table~\ref{tab:input_output_ablation}, input-output calibration generally outperforms the output-only variant across TAFAS, PETSA, and FAC. This suggests that output-side calibration alone may not be sufficient in many test-time adaptation settings, even though it is a simpler design choice. Importantly, the additional parameter cost of input-side calibration is modest, especially for FAC.

Table~\ref{tab:input_output_param_count} reports the corresponding trainable parameter counts. For each dataset and method, the parameter difference between the input-output and output-only variants corresponds to the input adapter. Since the parameter count of the input adapter is determined by the look-back window rather than the prediction horizon, its relative cost becomes smaller as $H$ increases.

\begingroup
\scriptsize
\setlength{\tabcolsep}{3pt}
\renewcommand{\arraystretch}{1.45}
\setlength{\LTpost}{10pt}

\begin{longtable}{cc|cc|cc|cc||cc|cc|cc}
\caption{
Input-output versus output-only calibration across source forecasters.
Lower MSE is highlighted within each input-output/output-only pair.
}
\label{tab:input_output_ablation}\\

\hline
Dataset & Horizon
& \multicolumn{6}{c||}{DLinear}
& \multicolumn{6}{c}{FreTS} \\
\cline{3-14}
& 
& \multicolumn{2}{c|}{TAFAS}
& \multicolumn{2}{c|}{PETSA}
& \multicolumn{2}{c||}{FAC}
& \multicolumn{2}{c|}{TAFAS}
& \multicolumn{2}{c|}{PETSA}
& \multicolumn{2}{c}{FAC} \\
\cline{3-14}
& 
& In+Out & Out
& In+Out & Out
& In+Out & Out
& In+Out & Out
& In+Out & Out
& In+Out & Out \\
\hline
\endfirsthead

\hline
Dataset & Horizon
& \multicolumn{6}{c||}{DLinear}
& \multicolumn{6}{c}{FreTS} \\
\cline{3-14}
& 
& \multicolumn{2}{c|}{TAFAS}
& \multicolumn{2}{c|}{PETSA}
& \multicolumn{2}{c||}{FAC}
& \multicolumn{2}{c|}{TAFAS}
& \multicolumn{2}{c|}{PETSA}
& \multicolumn{2}{c}{FAC} \\
\cline{3-14}
& 
& In+Out & Out
& In+Out & Out
& In+Out & Out
& In+Out & Out
& In+Out & Out
& In+Out & Out \\
\hline
\endhead

\hline
\endfoot

ETTh1 & 96
& \best{0.4617} & 0.4626
& \best{0.4613} & 0.4616
& \best{0.4554} & 0.4558
& \best{0.4391} & \best{0.4391}
& \best{0.4383} & \best{0.4383}
& \best{0.4357} & 0.4358 \\

& 192
& \best{0.5112} & 0.5115
& \best{0.5123} & 0.5128
& 0.5089 & \best{0.5071}
& \best{0.4943} & 0.4945
& \best{0.4960} & 0.4961
& 0.4934 & \best{0.4922} \\

& 336
& \best{0.5595} & 0.5600
& 0.5626 & \best{0.5622}
& 0.5582 & \best{0.5555}
& \best{0.5483} & 0.5485
& \best{0.5530} & \best{0.5530}
& 0.5498 & \best{0.5468} \\

& 720
& \best{0.6926} & 0.6970
& \best{0.6938} & 0.7003
& \best{0.6891} & 0.6918
& 0.6857 & \best{0.6855}
& \best{0.6908} & 0.6937
& 0.7067 & \best{0.7028} \\
\hline

ETTh2 & 96
& \best{0.2308} & 0.2315
& \best{0.2312} & 0.2316
& \best{0.2288} & 0.2302
& \best{0.2363} & 0.2366
& \best{0.2366} & 0.2370
& \best{0.2338} & 0.2341 \\

& 192
& 0.2830 & \best{0.2818}
& \best{0.2820} & 0.2834
& \best{0.2819} & 0.2832
& \best{0.2830} & 0.2831
& \best{0.2844} & 0.2846
& 0.2814 & \best{0.2810} \\

& 336
& \best{0.3175} & 0.3180
& \best{0.3190} & 0.3202
& \best{0.3144} & 0.3165
& \best{0.3195} & \best{0.3195}
& \best{0.3250} & 0.3252
& \best{0.3172} & 0.3180 \\

& 720
& \best{0.3825} & 0.3836
& \best{0.3884} & 0.3909
& \best{0.3825} & 0.3868
& \best{0.3814} & \best{0.3814}
& \best{0.3968} & 0.3971
& \best{0.3788} & 0.3862 \\
\hline

ETTm1 & 96
& \best{0.3504} & 0.3528
& \best{0.3541} & 0.3580
& \best{0.3465} & 0.3483
& \best{0.3559} & 0.3591
& \best{0.3572} & 0.3593
& \best{0.3517} & 0.3557 \\

& 192
& \best{0.4165} & 0.4172
& \best{0.4192} & 0.4226
& \best{0.4139} & 0.4160
& \best{0.4191} & 0.4200
& \best{0.4189} & 0.4201
& \best{0.4183} & 0.4187 \\

& 336
& \best{0.4801} & 0.4805
& \best{0.4831} & 0.4863
& \best{0.4791} & 0.4832
& \best{0.4810} & 0.4816
& \best{0.4793} & 0.4802
& 0.4836 & \best{0.4826} \\

& 720
& \best{0.5521} & 0.5526
& \best{0.5600} & 0.5618
& \best{0.5494} & 0.5517
& \best{0.5472} & 0.5475
& \best{0.5523} & 0.5526
& 0.5533 & \best{0.5496} \\
\hline

ETTm2 & 96
& \best{0.1581} & 0.1586
& \best{0.1583} & 0.1587
& \best{0.1560} & 0.1569
& \best{0.1566} & 0.1571
& \best{0.1569} & 0.1575
& \best{0.1563} & 0.1565 \\

& 192
& \best{0.1917} & 0.1919
& \best{0.1920} & 0.1922
& \best{0.1894} & 0.1901
& \best{0.1901} & 0.1904
& \best{0.1908} & 0.1912
& \best{0.1901} & 0.1903 \\

& 336
& \best{0.2299} & 0.2301
& \best{0.2303} & 0.2304
& 0.2286 & \best{0.2285}
& \best{0.2291} & 0.2293
& \best{0.2312} & 0.2314
& 0.2294 & \best{0.2287} \\

& 720
& 0.2993 & \best{0.2987}
& 0.2984 & \best{0.2968}
& 0.2984 & \best{0.2980}
& \best{0.2924} & 0.2926
& \best{0.2983} & 0.2984
& 0.2955 & \best{0.2934} \\
\hline

Weather & 96
& \best{0.1803} & 0.1839
& \best{0.1812} & 0.1876
& \best{0.1797} & 0.1881
& \best{0.1757} & 0.1787
& \best{0.1795} & 0.1814
& \best{0.1678} & 0.1764 \\

& 192
& \best{0.2265} & 0.2282
& \best{0.2308} & 0.2313
& \best{0.2231} & 0.2298
& \best{0.2167} & 0.2186
& \best{0.2229} & 0.2240
& \best{0.2114} & 0.2175 \\

& 336
& \best{0.2726} & 0.2737
& \best{0.2776} & 0.2778
& \best{0.2717} & 0.2777
& \best{0.2661} & 0.2669
& \best{0.2720} & \best{0.2720}
& \best{0.2621} & 0.2661 \\

& 720
& \best{0.3509} & 0.3542
& \best{0.3557} & 0.3565
& \best{0.3426} & 0.3482
& \best{0.3423} & 0.3426
& \best{0.3550} & 0.3554
& \best{0.3376} & 0.3392 \\
\hline

Exchange & 96
& \best{0.0875} & 0.0892
& \best{0.0898} & 0.0904
& \best{0.0875} & 0.0887
& \best{0.0796} & 0.0809
& \best{0.0825} & 0.0827
& \best{0.0810} & 0.0828 \\

& 192
& \best{0.1732} & 0.1758
& \best{0.1787} & 0.1802
& 0.1758 & \best{0.1740}
& \best{0.1654} & 0.1673
& \best{0.1729} & 0.1731
& 0.1667 & \best{0.1635} \\

& 336
& \best{0.3052} & 0.3083
& \best{0.3154} & 0.3188
& \best{0.2985} & 0.3125
& \best{0.3180} & 0.3197
& \best{0.3223} & 0.3230
& 0.3166 & \best{0.3017} \\
\hline

\end{longtable}

\renewcommand{\arraystretch}{1.55}
\makebox[\textwidth][c]{%
\resizebox{\textwidth}{!}{
\begin{tabular}{cc|cc|cc|cc||cc|cc|cc||cc|cc|cc}
\hline
Dataset & Horizon
& \multicolumn{6}{c||}{iTransformer}
& \multicolumn{6}{c||}{OLS}
& \multicolumn{6}{c}{PatchTST} \\
\cline{3-20}
& 
& \multicolumn{2}{c|}{TAFAS}
& \multicolumn{2}{c|}{PETSA}
& \multicolumn{2}{c||}{FAC}
& \multicolumn{2}{c|}{TAFAS}
& \multicolumn{2}{c|}{PETSA}
& \multicolumn{2}{c||}{FAC}
& \multicolumn{2}{c|}{TAFAS (PatchTST)}
& \multicolumn{2}{c|}{PETSA}
& \multicolumn{2}{c}{FAC} \\
\cline{3-20}
& 
& In+Out & Out
& In+Out & Out
& In+Out & Out
& In+Out & Out
& In+Out & Out
& In+Out & Out
& In+Out & Out
& In+Out & Out
& In+Out & Out \\
\hline

ETTh1 & 96
& 0.4421 & \best{0.4420}
& \best{0.4394} & 0.4398
& 0.4400 & \best{0.4378}
& \best{0.4428} & 0.4454
& \best{0.4449} & 0.4459
& \best{0.4382} & 0.4390
& 0.4270 & \best{0.4260}
& 0.4255 & \best{0.4253}
& 0.4241 & \best{0.4238} \\

& 192
& 0.5020 & \best{0.5016}
& \best{0.5014} & 0.5017
& 0.5010 & \best{0.4987}
& \best{0.4929} & 0.4951
& \best{0.5007} & 0.5016
& \best{0.4921} & 0.4926
& 0.4818 & \best{0.4811}
& \best{0.4818} & \best{0.4818}
& 0.4797 & \best{0.4790} \\

& 336
& \best{0.5624} & 0.5625
& 0.5641 & \best{0.5639}
& 0.5587 & \best{0.5561}
& \best{0.5427} & 0.5443
& \best{0.5488} & 0.5492
& 0.5422 & \best{0.5405}
& 0.5506 & \best{0.5505}
& \best{0.5570} & 0.5572
& 0.5525 & \best{0.5495} \\

& 720
& \best{0.6755} & 0.6772
& \best{0.6763} & 0.6783
& \best{0.6687} & 0.6718
& \best{0.6704} & 0.6747
& \best{0.6805} & 0.6835
& \best{0.6780} & 0.6832
& 0.6807 & \best{0.6802}
& 0.7019 & \best{0.7012}
& 0.7096 & \best{0.6964} \\
\hline

ETTh2 & 96
& \best{0.2620} & 0.2623
& 0.2602 & \best{0.2593}
& \best{0.2561} & \best{0.2561}
& \best{0.2289} & 0.2296
& \best{0.2293} & 0.2299
& \best{0.2274} & 0.2280
& 0.2380 & \best{0.2378}
& \best{0.2383} & \best{0.2383}
& \best{0.2374} & \best{0.2374} \\

& 192
& \best{0.3004} & 0.3011
& \best{0.3005} & 0.3008
& 0.2970 & \best{0.2955}
& \best{0.2821} & 0.2826
& \best{0.2829} & 0.2833
& \best{0.2796} & 0.2804
& \best{0.2752} & 0.2753
& \best{0.2810} & 0.2817
& 0.2747 & \best{0.2737} \\

& 336
& 0.3290 & \best{0.3286}
& 0.3312 & \best{0.3310}
& 0.3284 & \best{0.3266}
& \best{0.3189} & 0.3201
& \best{0.3240} & 0.3245
& \best{0.3137} & 0.3163
& \best{0.3108} & \best{0.3108}
& \best{0.3190} & 0.3200
& \best{0.3079} & 0.3103 \\

& 720
& \best{0.3959} & 0.3961
& \best{0.4049} & 0.4059
& 0.3982 & \best{0.3952}
& \best{0.3859} & 0.3870
& \best{0.3927} & 0.3955
& \best{0.3854} & 0.3901
& \best{0.4065} & 0.4073
& \best{0.4203} & 0.4215
& \best{0.4002} & 0.4043 \\
\hline

ETTm1 & 96
& \best{0.3680} & 0.3753
& \best{0.3700} & 0.3750
& \best{0.3459} & 0.3619
& \best{0.3542} & 0.3565
& \best{0.3591} & 0.3601
& \best{0.3462} & 0.3489
& \best{0.3826} & 0.3904
& \best{0.3817} & 0.3872
& \best{0.3530} & 0.3804 \\

& 192
& \best{0.4189} & 0.4238
& \best{0.4189} & 0.4238
& \best{0.4087} & 0.4159
& \best{0.4184} & 0.4203
& \best{0.4252} & 0.4266
& \best{0.4140} & 0.4171
& 0.4370 & \best{0.4367}
& 0.4405 & \best{0.4403}
& \best{0.4213} & 0.4269 \\

& 336
& \best{0.4800} & 0.4833
& \best{0.4788} & 0.4822
& \best{0.4781} & 0.4820
& \best{0.4811} & 0.4827
& \best{0.4896} & 0.4909
& \best{0.4794} & 0.4847
& \best{0.4864} & 0.4867
& \best{0.4970} & 0.4975
& \best{0.4787} & 0.4805 \\

& 720
& \best{0.5579} & 0.5595
& \best{0.5615} & 0.5622
& \best{0.5460} & 0.5506
& \best{0.5515} & 0.5524
& \best{0.5594} & 0.5603
& \best{0.5497} & 0.5535
& 0.5413 & \best{0.5402}
& \best{0.5479} & 0.5485
& \best{0.5399} & 0.5420 \\
\hline

ETTm2 & 96
& \best{0.1640} & 0.1642
& \best{0.1640} & 0.1642
& 0.1613 & \best{0.1606}
& \best{0.1594} & 0.1595
& \best{0.1595} & \best{0.1595}
& \best{0.1563} & 0.1573
& 0.1569 & 0.1571
& \best{0.1569} & 0.1573
& \best{0.1558} & 0.1562 \\

& 192
& 0.2164 & \best{0.2163}
& \best{0.2157} & 0.2164
& \best{0.2067} & 0.2094
& \best{0.1924} & 0.1925
& \best{0.1925} & 0.1927
& \best{0.1897} & 0.1907
& \best{0.2058} & 0.2063
& \best{0.2079} & 0.2084
& \best{0.1976} & 0.2018 \\

& 336
& \best{0.2735} & 0.2758
& \best{0.2653} & 0.2748
& \best{0.2623} & 0.2700
& \best{0.2308} & 0.2309
& \best{0.2316} & 0.2318
& \best{0.2289} & 0.2293
& \best{0.2482} & \best{0.2482}
& \best{0.2473} & 0.2476
& \best{0.2422} & 0.2458 \\

& 720
& 0.3303 & \best{0.3295}
& \best{0.3263} & 0.3264
& 0.3290 & \best{0.3251}
& \best{0.2996} & \best{0.2996}
& \best{0.3003} & \best{0.3003}
& \best{0.2973} & 0.2985
& \best{0.3218} & 0.3223
& \best{0.3240} & 0.3242
& \best{0.3192} & 0.3196 \\
\hline

Weather & 96
& \best{0.1662} & 0.1674
& \best{0.1662} & 0.1686
& \best{0.1605} & 0.1635
& \best{0.1825} & 0.1855
& \best{0.1890} & 0.1914
& \best{0.1728} & 0.1843
& \best{0.1700} & 0.1713
& \best{0.1676} & 0.1721
& \best{0.1635} & 0.1682 \\

& 192
& \best{0.2129} & 0.2136
& \best{0.2165} & 0.2171
& \best{0.2103} & 0.2113
& \best{0.2238} & 0.2255
& \best{0.2293} & 0.2304
& \best{0.2171} & 0.2249
& \best{0.2119} & 0.2130
& \best{0.2120} & 0.2137
& \best{0.2050} & 0.2088 \\

& 336
& \best{0.2629} & 0.2632
& \best{0.2672} & 0.2675
& 0.2618 & \best{0.2599}
& \best{0.2729} & 0.2740
& \best{0.2785} & \best{0.2785}
& \best{0.2665} & 0.2721
& \best{0.2659} & 0.2667
& \best{0.2703} & 0.2716
& \best{0.2570} & 0.2614 \\

& 720
& \best{0.3418} & 0.3422
& \best{0.3470} & 0.3474
& \best{0.3357} & 0.3365
& \best{0.3467} & 0.3475
& \best{0.3598} & 0.3604
& \best{0.3401} & 0.3424
& 0.3356 & \best{0.3348}
& \best{0.3529} & 0.3543
& \best{0.3341} & 0.3357 \\
\hline

Exchange & 96
& \best{0.0873} & 0.0877
& \best{0.0881} & \best{0.0881}
& 0.0867 & \best{0.0863}
& \best{0.0799} & 0.0805
& \best{0.0813} & \best{0.0813}
& \best{0.0793} & 0.0794
& \best{0.0816} & 0.0825
& \best{0.0850} & 0.0853
& \best{0.0813} & 0.0821 \\

& 192
& \best{0.1800} & 0.1820
& \best{0.1820} & 0.1836
& 0.1773 & \best{0.1756}
& \best{0.1650} & 0.1669
& \best{0.1719} & 0.1722
& \best{0.1648} & 0.1677
& 0.1666 & \best{0.1659}
& \best{0.1765} & 0.1769
& \best{0.1692} & 0.1718 \\

& 336
& \best{0.3014} & 0.3023
& \best{0.3105} & 0.3110
& \best{0.2983} & 0.3026
& 0.2846 & \best{0.2823}
& \best{0.3058} & 0.3093
& \best{0.2924} & 0.3009
& 0.3048 & \best{0.2763}
& \best{0.3188} & 0.3261
& \best{0.3011} & 0.3149 \\
\hline

\end{tabular}
}}
\endgroup

\begin{table*}[!htbp]
\centering
\vspace{1.5em}  
\scriptsize
\begingroup
\renewcommand{\arraystretch}{1.35}
\setlength{\tabcolsep}{4pt}
\caption{
Trainable parameter counts for input-output and output-only calibration.
For each method, ``In+Out'' denotes the standard input-output variant and ``Out'' denotes the output-only variant.
The ETT datasets share the same parameter counts.
For TAFAS, the PatchTST-specific implementation is listed separately.
}
\label{tab:input_output_param_count}
\resizebox{\textwidth}{!}{
\begin{tabular}{cc|cc|cc|cc|cc}
\hline
\rule[-5pt]{0pt}{21pt}
\raisebox{4pt}[0pt][0pt]{\textbf{Dataset}}
& \raisebox{4pt}[0pt][0pt]{\textbf{$H$}}
& \multicolumn{2}{c|}{\shortstack{\textbf{TAFAS}\\[-1pt]\vphantom{\textbf{(PatchTST)}}}}
& \multicolumn{2}{c|}{\shortstack{\textbf{TAFAS}\\[-1pt]\textbf{(PatchTST)}}}
& \multicolumn{2}{c|}{\shortstack{\textbf{PETSA}\\[-1pt]\vphantom{\textbf{(PatchTST)}}}}
& \multicolumn{2}{c}{\shortstack{\textbf{FAC}\\[-1pt]\textbf{(Ours)}}} \\
\cline{3-10}
& 
& In+Out & Out
& In+Out & Out
& In+Out & Out
& In+Out & Out \\
\hline

\raisebox{-7pt}[0pt][0pt]{%
  \shortstack{%
    \textbf{ETT}\\[4pt]
    \raisebox{-1pt}[0pt][0pt]{\textbf{(h1/2, m1/2)}}%
  }%
}
& 96  
& 130,382 & 65,191
& 18,626 & 9,313
& 25,934 & 12,967
& \best{2,758} & \best{1,379} \\

& 192 
& 324,590 & 259,399
& 46,370 & 37,057
& 38,894 & 25,927
& \best{4,102} & \best{2,723} \\

& 336 
& 857,822 & 792,631
& 122,546 & 113,233
& 58,334 & 45,367
& \best{6,118} & \best{4,739} \\

& 720 
& 3,699,038 & 3,633,847
& 528,434 & 519,121
& 110,174 & 97,207
& \best{11,494} & \best{10,115} \\
\hline

\textbf{Weather}
& 96  
& 391,146 & 195,573
& 18,626 & 9,313
& 71,658 & 35,829
& \best{8,274} & \best{4,137} \\

& 192 
& 973,770 & 778,197
& 46,370 & 37,057
& 107,466 & 71,637
& \best{12,306} & \best{8,169} \\

& 336 
& 2,573,466 & 2,377,893
& 122,546 & 113,233
& 161,178 & 125,349
& \best{18,354} & \best{14,217} \\

& 720 
& 11,097,114 & 10,901,541
& 528,434 & 519,121
& 304,410 & 268,581
& \best{34,482} & \best{30,345} \\
\hline

\textbf{Exchange}
& 96  
& 149,008 & 74,504
& 18,626 & 9,313
& 29,200 & 14,600
& \best{3,152} & \best{1,576} \\

& 192 
& 370,960 & 296,456
& 46,370 & 37,057
& 43,792 & 29,192
& \best{4,688} & \best{3,112} \\

& 336 
& 980,368 & 905,864
& 122,546 & 113,233
& 65,680 & 51,080
& \best{6,992} & \best{5,416} \\
\hline

\end{tabular}
}
\endgroup
\end{table*}

\subsection{Runtime and Parameter Efficiency}
\label{app:runtime}

\begin{table*}[!htbp]
\centering
\caption{
Runtime and parameter comparison on Weather with the longest forecasting horizon $H=720$, a relatively demanding setting in our benchmark. Adaptation time includes calibration, loss computation, backward update, and period estimation, excluding the source forecaster forward pass.
Results are averaged over 10 repeated runs.
}
\label{tab:runtime_parameter_comparison}
\resizebox{\textwidth}{!}{
\begin{tabular}{cccccc}
\hline
Source Forecaster & Method & Trainable Parameters & MSE & Adaptation Time & Overall Runtime \\
\hline

DLinear & TAFAS & 11,097,114 & \second{0.3509} & \best{$3.60 \pm 0.25$ ms} & \best{$3.49 \pm 0.14$ s} \\
        & PETSA & \second{304,410} & 0.3557 & $6.88 \pm 0.63$ ms & $3.85 \pm 0.23$ s \\
        & FAC (Ours) & \best{34,482} & \best{0.3426} & \second{$4.25 \pm 0.41$ ms} & \second{$3.55 \pm 0.15$ s} \\
\hline

FreTS & TAFAS & 11,097,114 & \second{0.3423} & \best{$5.81 \pm 0.18$ ms} & \second{$7.57 \pm 0.18$ s} \\
      & PETSA & \second{304,410} & 0.3550 & $21.35 \pm 0.56$ ms & $8.38 \pm 0.24$ s \\
      & FAC (Ours) & \best{34,482} & \best{0.3376} & \second{$6.01 \pm 0.39$ ms} & \best{$7.27 \pm 0.24$ s} \\
\hline

iTransformer & TAFAS & 11,097,114 & \second{0.3418} & \best{$7.29 \pm 1.15$ ms} & \best{$5.56 \pm 0.12$ s} \\
             & PETSA & \second{304,410} & 0.3470 & $10.32 \pm 0.58$ ms & $6.67 \pm 1.50$ s \\
             & FAC (Ours) & \best{34,482} & \best{0.3357} & \second{$7.92 \pm 1.35$ ms} & \second{$5.56 \pm 0.22$ s} \\
\hline

OLS & TAFAS & 11,097,114 & \second{0.3467} & \best{$3.10 \pm 0.31$ ms} & \best{$3.67 \pm 0.29$ s} \\
    & PETSA & \second{304,410} & 0.3598 & $6.79 \pm 0.46$ ms & $4.07 \pm 0.15$ s \\
    & FAC (Ours) & \best{34,482} & \best{0.3401} & \second{$3.54 \pm 0.28$ ms} & \second{$3.79 \pm 0.23$ s} \\
\hline

PatchTST & TAFAS (PatchTST) & 528,434 & \second{0.3356} & \best{$2.97 \pm 0.12$ ms} & \best{$20.76 \pm 0.29$ s} \\
         & PETSA & \second{304,410} & 0.3529 & $95.70 \pm 1.98$ ms & $27.78 \pm 0.40$ s \\
         & FAC (Ours) & \best{34,482} & \best{0.3341} & \second{$18.16 \pm 0.29$ ms} & \second{$27.21 \pm 0.20$ s} \\
\hline

\end{tabular}
}
\end{table*}

\noindent\textit{Note.}
TAFAS and PETSA are evaluated under the same protocol as FAC, where adaptation uses only matured ground truth. For PatchTST, TAFAS follows its original backbone-specific implementation, which uses a lightweight single-channel calibration module and skips input calibration. For frequency-domain calibration methods, including FAC and our PETSA variant, we use multivariate calibration across variables.

Runtime efficiency is important in practice, but reducing the number of trainable parameters does not necessarily yield proportional wall-clock speedups. According to Table~\ref{tab:runtime_parameter_comparison}, under the same adaptation protocol, TAFAS generally achieves lower runtime overhead, whereas FAC introduces frequency awareness through a compact set of trainable parameters. Since frequency-domain adaptation involves FFT/iFFT operations and additional tensor transformations, FAC and PETSA may incur higher runtime overhead despite using fewer trainable parameters. Compared with PETSA, whose complex adaptation objective includes frequency-domain, distributional, and correlation-based terms, FAC directly parameterizes a lightweight frequency-domain calibration module. As a result, FAC achieves better forecasting accuracy with substantially fewer trainable parameters, while maintaining runtime close to TAFAS and lower than PETSA in most settings, except for PatchTST where TAFAS uses a particularly lightweight single-channel calibration module.

\subsection{Additional Frequency-Domain Correction Spectra}
\label{app:freq_spectra}

We provide additional frequency-domain correction spectra across datasets and forecasting horizons to complement the analysis in Figure~\ref{fig:FAC_freq}. Recall that these spectra are not intended to measure forecasting error. Instead, they characterize the magnitude and structure of the prediction changes induced by adaptation in the frequency domain.

For each adaptation method \(q \in \{\mathrm{FAC}, \mathrm{TAFAS}, \mathrm{PETSA}\}\), we first compute the correction for each collected prediction window. Specifically, we denote the correction as 
\begin{equation*}
\Delta \hat{\mathbf{Y}}^{q,[n]}
=
\hat{\mathbf{Y}}^{q,[n]}
-
\hat{\mathbf{Y}}^{[n]},
\qquad n=1,\ldots,N,
\end{equation*}
where \(\hat{\mathbf{Y}}^{[n]}\) denotes the prediction before adaptation and \(\hat{\mathbf{Y}}^{q,[n]}\) denotes the prediction after applying method \(q\) to the \(n\)-th prediction window.  We then concatenate the corrections from all prediction windows along the prediction-window dimension, yielding
\begin{equation*}
\Delta \hat{\mathbf{Y}}^{q}
\in
\mathbb{R}^{N \times H \times C},    
\end{equation*}
where \(N\) is the total number of prediction windows, \(H\) is the forecasting horizon, and \(C\) is the number of channels. Next, we apply rFFT along the forecasting-horizon dimension and take the magnitude as
\begin{equation*}
\mathbf{A}^{q}
=
\left|
\operatorname{rFFT}_{H}
\left(
\Delta \hat{\mathbf{Y}}^{q}
\right)
\right|
\in \mathbb{R}^{N \times F \times C},
\end{equation*}
where \(F=\lfloor H/2 \rfloor+1\) is the number of rFFT frequency components. The plotted curve is obtained by averaging over the prediction windows and channels as
\begin{equation*}
\bar{\mathbf{a}}^{q}
=
\operatorname{Avg}_{N,C}
\left(
\mathbf{A}^{q}_{f>0}
\right)
\in \mathbb{R}^{F-1}.
\end{equation*}
Here, \(f>0\) indicates that the DC component is omitted.

\begin{figure*}[!htbp]
    \centering
    \includegraphics[width=0.98\textwidth]{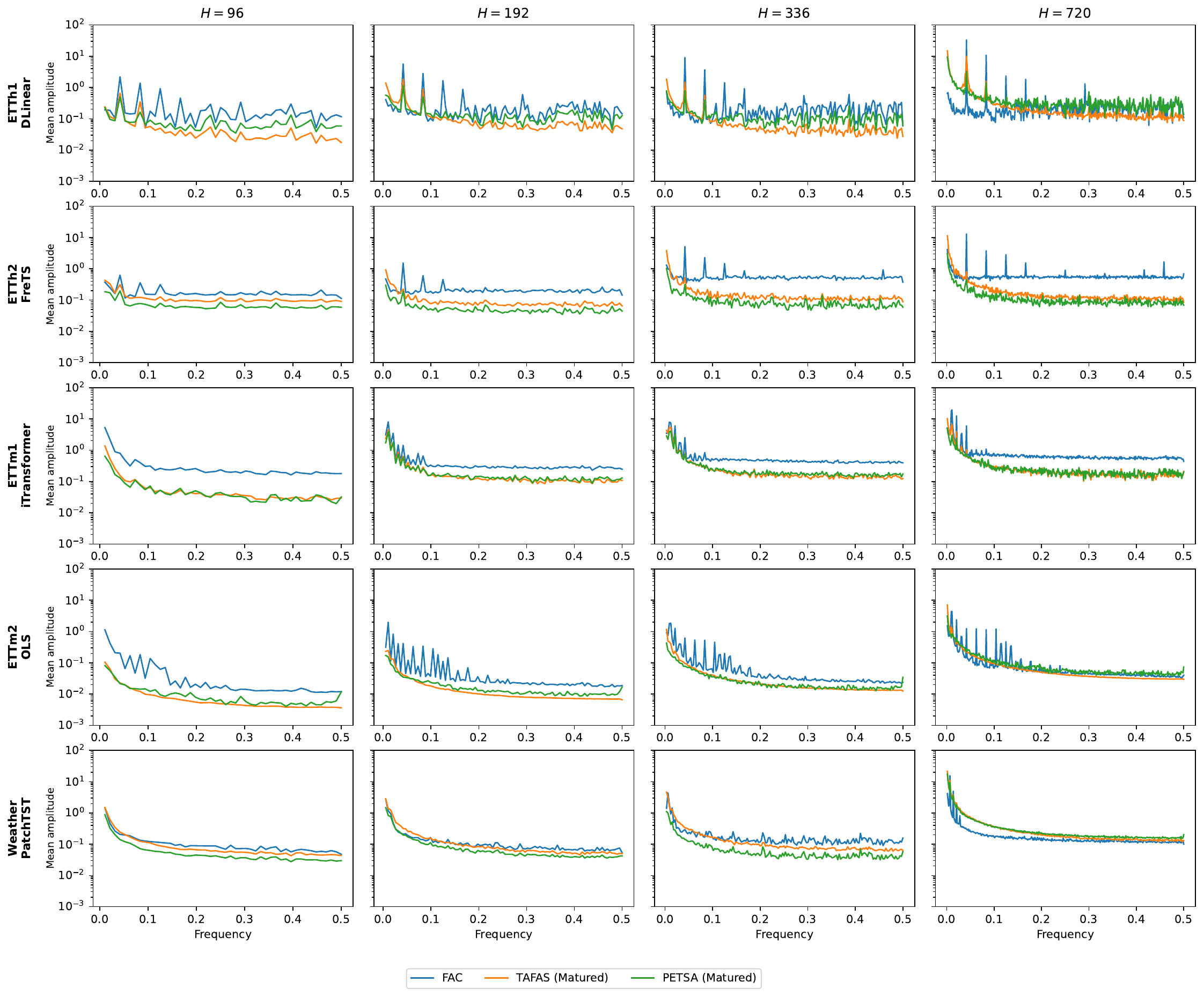}
    \caption{Additional frequency-domain correction spectra across forecasting horizons. For each adaptation method, we compute the correction, apply rFFT to this correction along the forecasting horizon, and plot the spectral magnitude averaged over prediction windows and channels. The DC component is omitted. Columns correspond to $H \in \{96,192,336,720\}$, and rows correspond to dataset--forecaster pairs ordered alphabetically by dataset name. The y-axis is shown on a logarithmic scale. We omit Exchange because no matured mini-batch is available under the $H=720$ setting in our protocol.}
    \label{fig:freq_spectra_all_horizons}
\end{figure*}

Figure~\ref{fig:freq_spectra_all_horizons} shows that the qualitative behavior observed in the main text is not limited to a single forecasting horizon. Across diverse settings, FAC produces more localized spectral corrections than TAFAS and PETSA, whose corrections often vary more smoothly across frequencies. At the same time, some cases, such as Weather with PatchTST, show smaller differences among the methods, indicating that the spectral behavior also depends on the source forecaster and dataset.

\end{document}